\documentclass[journal, 10pt]{IEEEtran}
\usepackage{amsmath,amsfonts}
\usepackage{algorithmic}
\usepackage{algorithm}
\usepackage{array}
\usepackage[caption=false,font=normalsize,labelfont=sf,textfont=sf]{subfig}
\usepackage{textcomp}
\usepackage{stfloats}
\usepackage{url}
\usepackage{verbatim}
\usepackage{graphicx}
\usepackage{cite}
\usepackage{authblk}
\usepackage{setspace}
\usepackage{booktabs}
\usepackage{pdflscape}
\usepackage[utf8]{inputenc}

\usepackage{mathtools}
\usepackage{xspace}
\usepackage{xcolor}
\usepackage{rotating}
\usepackage{longtable}
\usepackage{xr-hyper}
\usepackage{dirtree}

\date{}

\usepackage{mathtools}
\usepackage{xspace}
\usepackage{xcolor}
\usepackage{rotating}
\usepackage{longtable}

\def\yolof{\textsc{YOLO}v4\xspace}
\def\yolos{\textsc{YOLO}v7\xspace}
\def\yolon{\textsc{YOLO}v9\xspace}
\def\yoloe{\textsc{YOLO}11\xspace}
\def\yolot{\textsc{YOLO}v12\xspace}
\def\frcnn{Faster-\textsc{RCNN}\xspace} 
\def\yolo{\textsc{YOLO}\xspace}
\def\cnv2{\emph{CorNetv2}\xspace}

\def\smd{\emph{MSDD}\xspace}

\def\asift{\textsc{ASIFT}\xspace}	  
\def\uav{\textsc{UAV}\xspace}  
\def\uavs{\textsc{UAV}s\xspace}

\def\ffmpeg{\texttt{FFmpeg}\xspace}
\def\ymark{\texttt{Yolo\_mark}\xspace}

\def\map{\text{mAP}\xspace}

\def\yv12{\mbox{YOLOv12}\xspace}

\usepackage{hyperref}

\begin{document}

\title{\emph{Maize Seedling Detection Dataset (MSDD)}: A Curated High-Resolution RGB Dataset for Seedling Maize Detection and Benchmarking with \yolon, \yoloe, \yolot and \frcnn}

\author{Dewi Endah Kharismawati
        and Toni Kazic

\thanks{D. E. Kharismawati and T. Kazic are with the Department of Electrical Engineering and Computer Science, University of Missouri, Columbia, USA (e-mail: dek8v5@missouri.edu; kazict@missouri.edu).}
\thanks{Manuscript received 18 September 2025;}
}


\maketitle

\begin{abstract}
%

Accurate maize seedling detection is crucial for modern precision
agriculture, yet curated datasets for this purpose remain scarce.
To address this gap, we introduce \smd, a high-quality dataset of aerial
images designed for maize seedling stand counting, which has broad
applications in early-season crop monitoring, yield prediction, and
in-field management decision-making.
Stand counting helps determine how many plants have successfully
germinated, enabling farmers to make timely
adjustments such as replanting or modifying input applications.
Traditional stand counting methods are labor-intensive and error-prone,
whereas automated detection using computer vision offers significant improvements
in efficiency and accuracy. \smd comprises three classes --- single plant,
double plants, and triple plants --- to ensure precise stand counting.
This dataset includes aerial images captured under diverse conditions,
featuring different growth stages, planting
setups, soil colors, illuminations, camera angles, and planting
densities, making it robust for real-world applications.
Benchmarking results show that detection performs best during
the V4–V6 growth stages and under nadir camera views.
Among the evaluated models, \yoloe is the fastest, while \yolon provides
the highest accuracy for detecting single maize plants.
While single plant detection achieves high precision up to 0.984 and
recall up to 0.873, detecting double and triple plants remains challenging due
to their rarity and anomalous appearance, often caused by planting errors.
The imbalanced distribution of these categories contributes to reduced model
accuracy in identifying multiple plants per stand.
Despite these challenges, the fastest model, \yoloe, maintains efficient
inference with an average processing time of 35 milliseconds per image,
plus an additional 120 milliseconds for saving prediction bounding boxes and annotated images.
\smd provides a
critical foundation for developing robust detection models that enhance
stand counting precision, optimize resource allocation, and
and supports real-time decision-making throughout the field season.
This dataset marks a significant step toward the
automation of agricultural monitoring, contributing to
the advancement of precision agriculture.
%
%

%
%
%


\end{abstract}



\section{Introduction}
\label{sec:org53e8131}

Maize (\textit{Zea mays}), commonly known as corn, is one of the most widely
cultivated crops in the world, serving as a staple food source for millions
and playing a vital role in global food security.
Originating from Mesoamerica, maize was domesticated thousands of years ago
before spreading across the world through trade and colonization.
It became a dominant crop in the United States after its introduction to
early European settlers, evolving through centuries of breeding and
agricultural advancements \cite{revilla2022, galani2022, tanumihardjo2020}.
With a growing global population and the escalating impacts of climate
change, food production faces unprecedented challenges.
By 2050, global
food demand is expected to rise by 30-62\%, while the number of people
at risk of hunger could increase by up to 30\%. By 2080, an additional 170
million people may be at risk of hunger. Without innovative agricultural
solutions to improve crop resilience and yield, the world could face severe
food shortages \cite{dijk2021, schmidhuber2007, kumar2016, tilman2011, bodirsky2015, oecd2010}.
To meet this demand, researchers must continually adapt, focusing on
developing maize varieties that are more resilient to climate stressors
such as drought, extreme temperatures, and soil degradation.
Traditional plant phenotyping methods still rely heavily on manual
measurement, a laborious, time-consuming and error-prone process that
limits improvement efficiency, especially in large populations of plants
\cite{xiao2022,atefi2021}.
%
To address these challenges, advancements such as genomic selection, remote
sensing, and high-throughput digital phenotyping have emerged as promising
solutions to enhance genetic gain and accelerate crop
improvement \cite{gill2022, kharismawati2020, costa2019, aktar2020, kharismawati2025a, montesinos-lopez2016}.
Among these innovations, early-stage field assessments play a crucial role
in optimizing crop management and breeding selection.
Stand counting, early growth monitoring, weed detection, stress
identification, and yield prediction are essential for evaluating plant
robustness and making real-time informed decisions during the field season.
One critical phenotyping task is assessing germination success in new maize
lines to identify resilient genotypes.
This is especially important in genetic nurseries, where each row contains
distinct maize lines, as early stand counting offers valuable insights into
seedling germination and emergence rates.
Traditionally, stand counting is performed by manually walking the field and
counting seedlings, which is labor-intensive and often inaccurate
\cite{pathak2022, kharismawati2025, bullock1998}.
Automating this process with advanced computer
vision and deep learning techniques could significantly improve efficiency
and accuracy.
However, the development of such systems is hindered by the lack of a
publicly available, large-scale, and well-curated dataset for detecting
maize seedlings \cite{heider2025, david2020, chamorro-padia2024}.
Unlike conventional object detection datasets, agricultural datasets pose
unique challenges due to the uniformity of plants, their close spacing, and
the absence of clear separation between individual seedlings.
The difficulty is further compounded by variations in planting densities,
row configurations, and environmental factors across different agricultural
systems \cite{david2020, oh2020, ullah2024}.
Moreover, detection accuracy can be affected by variations in leaf shape,
size, and color, which differ significantly among maize seedling genotypes.
Additional challenges arise from different imaging perspectives, as
seedlings captured from nadir (top-down) and oblique (angled) views exhibit
different visual characteristics.
Soil color variations also influence
model performance, as seedlings appear differently depending on background
contrast.
Existing agricultural datasets are often limited in terms of genetic lines,
geographic location, and environmental variability, making it difficult to
develop robust and generalizable detection models
\cite{khun2021, david2020, varela2018}.
A common strategy in digital
phenotyping is to train deep learning models on datasets collected in
controlled environments.
However, models trained on such
idealized datasets often fail to perform well in real-world field
conditions due to domain shifts \cite{rodriguez-vazquez2023, wu2023a, arshad2024}.
Furthermore, the
process of collecting and labeling large-scale agricultural datasets is
labor-intensive and prone to errors, which limits the availability of
high-quality annotated data for seedling maize detection
\cite{xu2023, katari2024, cravero2022, waltz2025}.
Without access to a diverse and well-labeled dataset, research in this
field is hindered, slowing advancements in automated phenotyping and early stand
counting.
Providing such a dataset would not only accelerate
maize seedling detection research but also foster collaboration across
disciplines, including agronomy, computer vision, and machine
learning.
To address this gap, we present Maize Seedling Detection Dataset (\smd), a
large and diverse dataset for seedling maize detection aimed at improving
stand counting accuracy.
The dataset includes images captured at
different growth stages, from various camera angles, and across
multiple genotypes, soil colors, field setups, and planting
densities.
To enhance counting accuracy, \smd introduces
three classification categories: single plant, two plants, and three
plants.
This paper details the dataset's collection, labeling
methodology, and organization, providing a valuable resource for
researchers to develop and refine deep learning models for maize seedling
detection.

\section{Materials and Methods}
\label{sec:org69b7893}

\subsection{Maize Nurseries}
\label{sec:org398c59a}

Maize genetic nurseries were planted and imaged in 2019, 2020, 2021, and
2022 essentially as described in \cite{kharismawati2020}.
Briefly, our fields were planted by hand using a jab planter for
disease lesion mimic mutant and inbred lines, a Jang rotary push planter
for an elite line, and a two-row cone planter for machine planting in the
borders and adjacent research fields.
%
%
In all methods, small holes were made in the soil, seeds were dropped, and
the holes were covered with soil and firmed by foot pressure. 

The nursery
%
%
planted in
6.1 m rows spaced 0.91 m apart. In manually planted fields, ranges were
separated by 1.22 m unplanted alleys;
%
%
machine-planted fields had no
alleys.
While plant spacing within rows varied, our average spacing was
approximately 0.30 m between plants.
Maize growth stages were monitored using the ``leaf collar'' method, which
counts the number of leaves the plant has, starting at 1 with the
coleoptile \cite{nielsen2004}.
For our lines, the upper bound for plants
at V4 is approximately 12cm high; for V8, approximately 25cm;
and for V10, approximately 40cm.

\subsection{Equipment}
\label{sec:orgc613d33}

Data collection was conducted using a DJI Phantom 4 and DJI Mavic 2 Pro by
Da-Jiang Innovations, Shenzhen, China for RGB video imaging.  All flights
were flown manually using either
the Autopilot mobile app or  DJI Go 4.
Video was collected using both nadir and oblique camera angles, recorded at
24 and 30 frames \emph{per} second (fps).

Video frames were extracted using \href{https://www.ffmpeg.org}{\ffmpeg}
with different sampling rate
(refer to Tables \ref{train-data}, \ref{val-data}, \ref{test-data}).
Seedling labelling was conducted using
\href{https://github.com/AlexeyAB/Yolo_mark}{\ymark}.
All computations were performed on a Lambda Labs machine equipped with an
Intel Core i9-9920X CPU, two NVIDIA RTX 2080Ti GPUs, and 128 GB of
RAM.
Semi-automatic data labelling, benchmarking, and data analysis involved
\textsc{YOLO}v3, \yolof, \yolos, \yolon, \yoloe, \yolot, and \frcnn
\cite{redmon2018, bochkovskiy2020, wang2022, wang2024, ren2016, jocher2023, tian2025}.

\subsection{Data Collection}
\label{sec:org72efb36}

The \uavs were freely flown at speeds ranging from 0.8 to 3.2 km/h
%
%
with
three primary flight trajectories: serpentine forward-backward passes
parallel to the crop rows with lateral slides between passes; similar
passes perpendicular to the crop rows with slides between passes; and
perpendicular passes combined with rotations between
passes \cite{kharismawati2025a}.
Some flights were unconstrained, further diversifying the dataset by
introducing natural variations.
Data collection took place
%
%
at different times of
the day to account for varying light conditions and shadows, improving the
dataset's robustness.
To ensure high resolution imagery, data collection was conducted using
low altitude flights, ranging from approximately 1 to 18 meters above
ground level (AGL).
%
%
%
%
%
The vehicle was flown in relatively
light wind conditions, but occasionally experienced horizontal and vertical
displacements due to air currents.
At an altitude of 4.5 meters, the ground sampling distance was
approximately 0.5 cm per pixel.
The drone cameras were positioned in both nadir and oblique
angles, providing multiple perspectives that revealed variations in
seedling shape and morphology.

\subsection{Image Dataset}
\label{sec:org03878ae}

The dataset includes diverse imagery capturing various real-world
conditions in both research and production maize fields, ensuring
robustness for training and evaluation.
These images encompass a range of genetic backgrounds, including inbred
backgrounds, mutant lines, and elite hybrids.
The dataset spans growth stages from V2 to V10 and accounts for different
soil colors from rainfall, irrigation, and subsequent drainage.
Additionally, the dataset includes variations in plant orientations and
shapes influenced by genetic morphology, camera angles, occlusions due to
overlapping plants, uneven spacing, and camera parallax.
A summary of the videos used for training, validation, and testing is
provided in Tables \ref{train-data}, \ref{val-data}, and \ref{test-data}.
%

%
%
%
To ensure
rigorous model evaluation while minimizing data leakage and
overfitting, the dataset is structured into distinct training, validation,
and test sets.
The training dataset comprises images collected during the 2019, 2020, and
2021 field seasons, with the 2021 dataset including plots intentionally
planted with
%
%
double and triple clusters of seeds
%
%
to help balance the number of
objects between classes.
Validation data were also sampled from the 2019, 2020, and 2021 field
seasons but were selected from different videos to prevent overlap
with the training set while maintaining similar environmental conditions.
This ensures that the validation data remain
%
%
unseen during training while
still representing the same plant populations and fields.
The test dataset was collected during the 2022 field season
%
%
and consists of
entirely new fields and plants to evaluate the model's generalizability to
unseen data.
%


\begin{table*}
\centering
\caption{Training data for \smd.  Videos are identified by their unique id;
Objects are annotated either manually (denoted as $D_m$ in the annotation column),
semi-automatically ($D_s$),
or using homography projection ($D_h$).
Camera poses are nadir (n) or oblique (o).
Alt, approximate altitudes above ground level from vehicle telemetry in meters.
Stage, average growth stage of plants at imaging, estimated by inspecting rows.
Sampling is the sampling rate of the video in frames per second (fps).
A small sample of frames from videos 0573 and 0594 in $D_m$ were used in the initial training ($M_m$);
the entire videos were used in $D_s$.}
\begin{tabular}{|r|r|c|c|r|l|c|c|c|}
\hline
Year & VideoID       & Annotation   & Pose   &  Alt   & Stage   & Sampling   & No. Frames   & No. Patches \\
\hline
2019 &           0077 & $D_m$  & n      &       9.1 & V2--V7  & 1/5        &            76 & 152 \\
2019 &           0179 & $D_m$  & n      &       9.1 & V7      & 1/5        &             6 &  12 \\
2019 &           0218 & $D_m$  & n      &       6.1 & V2--V10 & 1/5        &             8 &  16 \\
2019 &           0219 & $D_m$  & n      &       6.1 & V2--V6  & 1/5        &             5 &  10 \\
2019 &           0288 & $D_m$  & n, o   & 1.5--15.2 & V2--V10 & 1/5        &             8 &  16 \\
2019 &           0289 & $D_m$  & n      &       6.1 & V4      & 1/5        &             8 &  16 \\
2019 &           0290 & $D_m$  & n      & 3.0--15.2 & V4--V10 & 1/5        &             6 &  12 \\
2019 &           0291 & $D_m$  & n      & 1.5--15.2 & V4--V10 & 1/5        &             7 &  14 \\
2019 &           0433 & $D_m$  & n      &      15.2 & V5      & 1/5        &             2 &   4 \\
2019 &           0433 & $D_m$  & o      &       4.6 & V5      & 1/5        &             5 &  10 \\
2019 &           0439 & $D_m$  & o      &      12.2 & V5      & 1/1        &             1 &   2 \\
2019 &           0450 & $D_m$  & n      & 1.5--12.2 & V5      & 1/5        &             6 &  12 \\
2019 &           0454 & $D_m$  & n      &       3.0 & V5      & 1/5        &             5 &  10 \\
2019 &           0457 & $D_m$  & n      &  1.5--6.1 & V5      & 1/5        &            11 &  22 \\
2019 &           0487 & $D_m$  & n      &       3.0 & V4--V6  & 1/5        &            15 &  30 \\
2020 &           0573 & $D_m$  & n      &       6.1 & V2--V8  & 1/5        &            21 &  42 \\
2020 &           0594 & $D_m$  & n      &       4.6 & V4--V6  & 1/5        &            14 &  28 \\
\hline
2020 &           0573 & $D_s$  & n      &       6.1 & V2--V8  & 1/2        &           145 & 290 \\
2020 &           0575 & $D_h$  & n      &      16.8 & V2--V8  & 1/2        &            47  & 94 \\
2020 &           0576 & $D_s$  & n      &       6.1 & V2--V8  & 1/2        &            87 & 174 \\
2020 &           0594 & $D_s$  & n      &       4.6 & V4      & 1/2        &           226 & 452 \\
2020 &           0603 & $D_s$  & o      &       6.1 & V6      & 1/2        &             1 &   2 \\
2020 &           0604 & $D_s$  & n      &       6.1 & V6      & 1/2        &            10 &  20 \\
2020 &           0605 & $D_s$  & n      &       7.6 & V6      & 1/2        &            10 &  20 \\
2020 &           0606 & $D_s$  & n      &       6.1 & V6      & 1/2        &            60 & 120 \\
2020 &           0614 & $D_s$  & n      &       4.6 & V6--V8  & 1/5        &             8  & 16 \\
2020 &           0615 & $D_s$  & o      &       7.6 & V6      & 1/2        &            37 &  74 \\
\hline
2021 &           0303 & $D_m$  & n      &       6.1 & V2--V4   & 1/9        &            27 &  54 \\
2021 &           0311 & $D_m$  & n      &       6.1 & V2--V4   & 1/9        &            19 &  38 \\
2021 &           0315 & $D_m$  & n      &       6.1 & V4--V8   & 1/9        &            14 &  28 \\
2021 &           0391 & $D_m$  & o      &       4.6 & V6--V10  & 1/9        &             4 &   8 \\
2021 &           0401 & $D_m$  & n      &       4.6 & V8--V12  & 1/9        &            10 &  20 \\
\hline 
\end{tabular}
\label{train-data}
\end{table*}

\begin{table*}
\centering
\caption{Validation data are semi-automatically marked ($D_{v,s}$)
and manually marked for the double and triple classes to match the actual
count in the field ($D_{v,m}$)}
\begin{tabular}{|r|r|c|c|r|l|c|c|c|}
\hline
Year & VideoID        & Annotation   & Pose   &  Alt   & Stage   & Sampling   & No. Frames   & No. Patches \\
\hline
2019 &           0076 & $D_{v,s}$  & n      &     9.1 & V2--V4  & 1/5        &             2  &  4 \\
2019 &           0078 & $D_{v,s}$  & n      &     7.6 & V4--V8  & 1/10       &             4  &  8 \\
2019 &           0080 & $D_{v,s}$  & o      &     9.1 & V4--V8  & 1/10       &             1  &  4 \\
2019 &           0081 & $D_{v,s}$  & n      &     6.1 & V6--V8  & 1/10       &             1  &  2 \\
2019 &           0083 & $D_{v,s}$  & o      &     9.1 & V6--V8  & 1/10       &             1  &  6 \\
2019 &           0463 & $D_{v,s}$  & n      &     3.0 & V4--V6  & 1/10       &             3  &  6 \\
2019 &           0481 & $D_{v,s}$  & n      &     1.5 & V4--V6  & 1/10       &             4  &  8 \\
2019 &           0499 & $D_{v,s}$  & n      &     3.0 & V4--V6  & 1/10       &             3  &  6 \\
2020 &           0577 & $D_{v,s}$  & n      &     4.6 & V2--V8  & 1/2        &            64 & 128 \\
2020 &           0579 & $D_{v,s}$  & n      &     4.6 & V2--V8  & 1/5        &            14  & 28 \\
2020 &           0580 & $D_{v,s}$  & n      &     4.6 & V2--V8  & 1/5        &             8  & 16 \\
2020 &           0581 & $D_{v,s}$  & n      &     3.0 & V2--V8  & 1/10       &             2  &  3 \\
2020 &           0587 & $D_{v,s}$  & n      &     4.6 & V4--V8  & 1/10       &             5  &  9 \\
2020 &           0592 & $D_{v,s}$  & n      &     7.6 & V4      & 1/2        &            43  & 86 \\
2020 &           0593 & $D_{v,s}$  & n      &     7.6 & V4      & 1/10       &             4  &  8 \\
2020 &           0596 & $D_{v,s}$  & n      &     4.6 & V2--V8  & 1/10       &             3  &  6 \\
2020 &           0597 & $D_{v,s}$  & n      &     4.6 & V2--V8  & 1/2        &            36  & 72 \\
2020 &           0607 & $D_{v,s}$  & n      &     6.1 & V6      & 1/2        &            14  & 28 \\
2020 &           0611 & $D_{v,s}$  & n      &     3.0 & V6--V8  & 1/10       &             1  &  2 \\
2020 &           0613 & $D_{v,s}$  & n      &     3.0 & V6--V8  & 1/10       &             2  &  4 \\
2020 &           0622 & $D_{v,s}$  & n      &     6.1 & V4--V7  & 1/10       &             1  &  1 \\
2020 &           0623 & $D_{v,s}$  & n      &     6.1 & V4--V6  & 1/5        &             8  & 16 \\
2020 &           0633 & $D_{v,s}$  & n      &     4.6 & V6--V10 & 1/5        &             6  & 11 \\
2020 &           0635 & $D_{v,s}$  & n      &     9.1 & V3--V6  & 1/10       &             4  &  8 \\
2020 &           0636 & $D_{v,s}$  & n      &     6.1 & V6--V10 & 1/10       &             3  &  6 \\
\hline
2021 &           0304 & $D_{v,m}$  & n      &     4.6 & V2--V4   & 1/9        &             3  &  6 \\
2021 &           0359 & $D_{v,m}$  & n      &     4.6 & V4--V8   & 1/9        &             9  & 18 \\
2021 &           0389 & $D_{v,m}$  & o      &     6.1 & V6--V10  & 1/9        &             3  &  6 \\
2021 &           0402 & $D_{v,m}$  & n      &     6.1 & V6--V10  & 1/9        &             11 & 22 \\
\hline
\end{tabular}
\label{val-data}
\end{table*}

\begin{table*}
\centering
\caption{
Test data were semi-automatically marked with \yolos, then validated individually manually by D.E.K. 
}
\begin{tabular}{|r|r|c|c|r|l|c|c|c|}
\hline
Year & VideoID       & Annotation   & Pose   &  Alt   & Stage   & Sampling   & No. Frames   & No. Patches \\
\hline
2022 &           0827 & $D_{t,s}$  & n      &     6.1  & V2--V4  & 1/20       &            12 & 24 \\
2022 &           0876 & $D_{t,s}$  & n      &    19.8  & V2--V4  & 1/20       &            14 & 28 \\
2022 &     0941--0943 & $D_{t,s}$  & n      &     9.1  & V2--V4  & 1/20       &            34 & 68 \\
2022 &     0953--0955 & $D_{t,s}$  & n      &     7.6  & V2--V4  & 1/20       &            40 & 80 \\
\hline
2022 &           0870 & $D_{t,s}$  & n      &     9.1  & V4--V6  & 1/20       &            15 & 30 \\
2022 &           0873 & $D_{t,s}$  & n      &    13.7  & V4--V6  & 1/15       &             7 & 14 \\
2022 &     0001--0002 & $D_{t,s}$  & o      &     3.0  & V4--V6  & 1/20       &            18 & 36 \\
2022 &     0993--0994 & $D_{t,s}$  & n      &     9.1  & V4--V6  & 1/20       &            30 & 60 \\
2022 &     0997--0998 & $D_{t,s}$  & n      &    10.7  & V4--V6  & 1/20       &            30 & 60 \\
2022 &           0021 & $D_{t,s}$  & o      &     4.6  & V4--V6  & 1/10       &            16 & 32 \\
2022 &     0023--0024 & $D_{t,s}$  & n      &     9.1  & V4--V6  & 1/15       &            40 & 80 \\
\hline
2022 &           0939 & $D_{t,s}$  & n      &    12.2  & V6--V12 & 1/15       &            13 & 26 \\
2022 &           0984 & $D_{t,s}$  & n      &    12.2  & V6--V10 & 1/20       &            15 & 30 \\
2022 &     0054--0056 & $D_{t,s}$  & n      &    12.2  & V6--V10 & 1/20       &            30 & 60 \\
2022 &     0061--0062 & $D_{t,s}$  & n      &    12.2  & V6--V10 & 1/10       &            48 & 96 \\
2022 &           0068 & $D_{t,s}$  & o      &     3.0  & V6--V10 & 1/10       &            21 & 42 \\
2022 &           0092 & $D_{t,s}$  & n      &    12.2  & V6--V10 & 1/15       &            20 & 40 \\
\hline
\end{tabular}
\label{test-data}
\end{table*}

\subsection{Data Preprocessing}
\label{sec:org91a8d64}

\begin{figure*}[!t]
    \centering
    \subfloat[DJI Phantom 4]{%
        \includegraphics[width=0.48\linewidth]{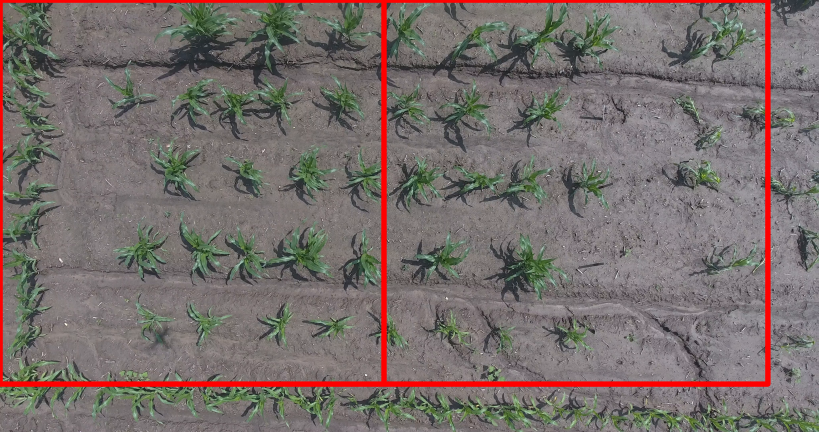}%
        \label{barbara-patch}
    }
    \hfil
    \subfloat[DJI Mavic 2 Pro]{%
        \includegraphics[width=0.48\linewidth]{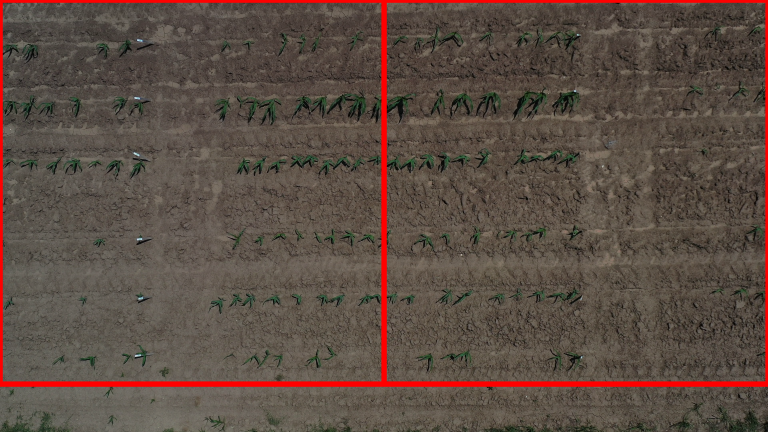}%
        \label{grace-patch}
    }
    \caption{Sample patchification throughout the dataset. The areas outside the red boxes are discarded.}
    \label{patch}
\end{figure*}

To ensure diversity in the dataset, videos were carefully selected to
include variations in maize lines, growth stages, camera angles, soil
colors, wind conditions, and flight trajectories.
Each video was then individually processed based on its duration to ensure
sufficient representation in the dataset.
The frame extraction rate varied between videos (see Tables \ref{train-data}, \ref{val-data}, and \ref{test-data}), but
frames were sampled sparsely to minimize duplication and avoid marking the
same plants multiple times.
Extracted frames followed a standardized naming convention:
\texttt{YY\_MM\_DD\_videoID\_FrameXXXXXX.png}, where \texttt{YY\_MM\_DD} denotes
the data collection date, \texttt{videoID} represents the three-digit video
ID, and \texttt{XXXXXX} corresponds to the six-digit frame sequence
number.
All images were stored in PNG format to preserve quality.
Once the frames were extracted, their dimensions varied depending on the
\uav.
%
Frames from the DJI Phantom 4 had a resolution of \(4096 \times 2160\)
pixels, while those from the DJI Mavic 2 Pro were \(3940 \times 2160\) pixels.
Since most deep learning architectures require square input dimensions,
each frame was divided into two non-overlapping \(1920 \times 1920\) fragments.
The fragmented frames followed a consistent naming convention:
\texttt{YY\_MM\_DD\_videoID\_FrameXXXXXX\_fragZ.png}, where \texttt{Z}
represents the fragment sequence number, ensuring traceability to their
original frames. Figure \ref{patch} shows example of patchification on frames captured
with the Phantom and Mavic.

\subsection{Data Labeling}
\label{sec:org74a9645}

In supervised learning for object detection, labeled data is essential for
training a network to recognize objects accurately.
We adopted a rinse-and-repeat approach to incrementally refine and scale up
our dataset.
The initial step involved manually annotating each maize plant in selected
frames by drawing bounding boxes using
\href{https://github.com/AlexeyAB/Yolo_mark}{\ymark}, forming dataset
\(D_m\).
Once an object is marked with a bounding box, \ymark automatically
generates a corresponding \texttt{.txt} file with the same filename as the
image.
Each bounding box is recorded as a separate row in the .txt file,
storing the object class, normalized coordinates of the bounding box center
(\(x, y\)), and its width and height (\(w, h\)) relative to the image
dimensions. 
For maize plants, the object class represented the number of plants within
the bounding box: single (one plant), double (two plants), and triple
(three plants).
The initial bounding boxes were relatively larger compared to the plants,
so subsequent annotations were refined to fit them more precisely.
Manual annotation proved to be time-consuming, with 220 frames requiring
two weeks for a single person to complete.
To improve efficiency, we implemented a semi-automatic labeling approach.
We fine-tuned the \yolof model using \(D_m\) over 12,000 iterations to
produce a new model, \(M_m\) \cite{bochkovskiy2020}.
This model was then used to label an additional 780 frames.
However, \(M_m\) failed to detect approximately 30\% of maize plants, mainly
younger ones at the V2--V4 growth stages, especially from altitudes above 15
meters.
To address this limitation, we incorporated sequences of frames collected
at higher altitudes and manually annotated those that captured entire field
ranges and rows.
Using homography matrices computed for mosaicking
\cite{kharismawati2020, kharismawati2025a}, bounding boxes from these frames
were projected onto adjacent frames that only partially covered the same
rows.
Any incomplete bounding boxes at the frame edges were discarded. Finally,
all labeled frames were manually reviewed to correct missing annotations
and remove false positives, forming dataset \(D_s\).

\subsection{Training}
\label{sec:org0dbb4e8}
\label{train-method}
To benchmark object detection performance on \smd, we trained \yolon,
\yoloe, \yolot, and \frcnn.
\yolon model \texttt{yolo9c} utilizes cross-scale feature fusion, dynamic label
assignment, and a compound backbone architecture to enhance detection
robustness across varied object sizes and occlusion levels \cite{wang2024}.
\yoloe model \texttt{yolo11x}, developed by Ultralytics, features an enhanced
backbone and neck architecture, refined training pipeline, and anchor-free
prediction for improved accuracy, speed, and efficiency across real-time
aerial imagery tasks \cite{jocher2023}.
\yolot model \texttt{yolo12x} leverages its enhanced feature aggregation,
multi-scale detection, and transformer-based attention mechanisms for
improved small-object recognition \cite{tian2025}.
\frcnn was trained using the Detectron2 framework with the
\texttt{faster\_rcnn\_R\_50\_FPN\_3x.yaml} configuration, which employs a ResNet-50
backbone, Feature Pyramid Network (FPN), and Region Proposal Network (RPN)
for accurate multi-scale object detection in complex field scenarios
\cite{ren2016}.
Training was conducted on our Lambda Lab machine using Python 3.11 and
PyTorch 2.2.2+cu121 with two NVIDIA GeForce RTX 2080 Ti GPUs.
Images underwent preprocessing to enhance generalization, including mosaic
augmentation, resizing, normalization, and additional augmentations such as
random flipping, brightness adjustments, rotation, and scaling
\cite{wang2024, jocher2023, tian2025, ren2016}.
The batch sizes were set to 8, 14, 18, and 4, for \yolon,
\yoloe, \yolot, and \frcnn respectively --  the maximum our
machine could handle.
%
%
Training was initially scheduled for 1000 epochs but
was stopped early at 513, 157, and 296 epochs due to convergence, except
for \frcnn, which was trained for the full 1000 epochs.
An initial learning rate of 0.001 was used, combined with a learning rate
decay schedule and a linear warmup during the first three epochs. This
adaptive learning rate strategy was applied to stabilize training and
prevent overfitting.

\subsection{Evaluation}
\label{sec:org3e7d373}

To provide an initial benchmark for object detection performance on our
dataset, we evaluated the model using standard object detection metrics,
including Bounding Box Accuracy, Intersection over Union (IoU), Mean
Average Precision (mAP), and Confusion Matrix Analysis.
These metrics establish a baseline for future improvements and facilitate
comparisons with alternative models trained on the dataset.
\textbf{Bounding Box Accuracy and IoU.} Bounding box accuracy was assessed using
Intersection over Union (IoU), which measures the overlap between the
predicted bounding box and the ground truth. IoU was computed as:
\begin{equation}
IoU = \frac{|B_p \cap B_g|}{|B_p \cup B_g|}.
\label{iou}
\end{equation}
where \(B_p\) represents the predicted bounding box and \(B_g\) represents the
ground truth bounding box.
A detection was considered correct if the IoU exceeded a predefined
threshold.
We report results for IoU
at a threshold of
0.5 as well as IoU thresholds between 0.50 to 0.95,
which evaluates model robustness across varying IoU thresholds.
These measures provide insight into how well the dataset supports accurate
localization of maize plants while avoiding overfitting.
\textbf{Mean Average Precision (mAP).}
As an initial benchmark, we computed mean Average Precision (mAP), a
standard metric for object detection that aggregates precision-recall
performance across different IoU thresholds. The mAP was calculated as:
\begin{equation}
mAP = \frac{1}{N} \sum_{i=1}^{N} AP_i.
\label{map}
\end{equation}
where \(AP_i\) represents the area under the precision-recall curve for class
\(i\), and \(N\) is the number of object classes.
We report mAP at an IoU threshold of 0.50 (mAP@0.50) as well as mAP
averaged over multiple IoU thresholds ranging from 0.50 to 0.95 in
increments of 0.05, to evaluate model robustness across varying overlap
criteria.

\textbf{Confusion Matrix.}
Since the dataset differentiates between single, double, and triple maize
plants per bounding box, we further analyzed detection performance using a
confusion matrix: correctly detected plants (True Positives); incorrectly
detected plants (False Positives); and missed detections (False Negatives).
From this, we derived Precision, Recall and F1, which indicate how
well the dataset supports robust plant classification:
\begin{equation}
\text{Precision} = \frac{TP}{TP + FP};
\label{precision}
\end{equation}

\begin{equation}
\text{Recall} = \frac{TP}{TP + FN}; \ \ \text{and} 
\label{recall}
\end{equation}

\begin{equation}
F1 = 2 \times \frac{\text{Precision} \times \text{Recall}}{\text{Precision} + \text{Recall}}.
\label{fscore}
\end{equation}
%

\textbf{Variations within the Dataset and Detection Performance.}
%
%
%
To evaluate how well the dataset supports robust detection, we analyzed
%
%
model performance across various conditions. 
Detection accuracy was assessed across different growth
stages (V2--V10) to determine how plant development affects detection
accuracy.
We also examined the impact of flight altitude (1m--18m) on
detectability, as higher altitudes reduces plant resolution.
Additionally, camera angles (nadir vs. oblique) were
considered to understand how variations in perspective influence the
detection of plant morphology.
%
%
%
Finally, we analyzed the effects of
environmental conditions, such as soil color, wind, and shadows, to assess
their influence on detection reliability.
By benchmarking performance
across these conditions, we establish a reference point for future
improvements in both dataset diversity and model optimization.

\section{Results}
\label{sec:org4343205}

The
%
%
paper presents a comprehensive maize seedling dataset designed to
support the development of robust detection models.
The dataset captures diverse growth stages and planting methods, including
separated, touching, and clustered maize seedlings.
To enhance its applicability, the dataset encompasses variations in maize
lines, altitudes, trajectories, camera poses, and soil colors.
This diversity ensures that models trained on it can generalize across
different agricultural conditions.
However, very young seedlings (smaller than V3) generate weak visual
signals, making their detection challenging, while overly large plants
(bigger than V8) frequently touch and occlude one another, making them
clustered and harder to detect. 
The following sections detail the dataset, labeling challenges, detection
performance, and model generalization.

\subsection{Dataset Overview}
\label{sec:org926c565}

\begin{figure}[!t]
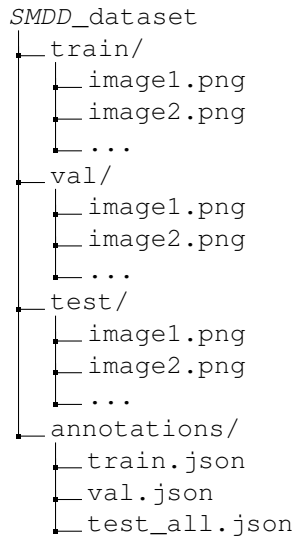

\centering

\begin{minipage}[t]{0.45\linewidth}
\yolo format\\[0.5ex]
\renewcommand*\DTstyle{\ttfamily}
\dirtree{%
.1 \emph{SMDD}\_dataset.
  .2 training/.
    .3 images/.
    .3 labels/.
  .2 validation/.
    .3 images/.
    .3 labels/.
  .2 test/.
    .3 images/.
    .3 labels/.
  .2 training.txt.
  .2 validation.txt.
  .2 test\_all.txt.
  .2 dmc\_data.yaml.
  .2 obj.names.
}
\label{yol-formt}
\end{minipage}
\hfil
\begin{minipage}[t]{0.45\linewidth}
\frcnn format\\[0.5ex]
\renewcommand*\DTstyle{\ttfamily}
\dirtree{%
.1 \emph{SMDD}\_dataset.
  .2 train/.
    .3 image1.png.
    .3 image2.png.
    .3 \dots.
  .2 val/.
    .3 image1.png.
    .3 image2.png.
    .3 \dots.
  .2 test/.
    .3 image1.png.
    .3 image2.png.
    .3 \dots.
  .2 annotations/.
    .3 train.json.
    .3 val.json.
    .3 test\_all.json.
}
\label{frcnn-formt}
\end{minipage}

\caption{
Directory structure for the dataset.
\yolo uses separate folders for images and labels,
while \frcnn (COCO format) centralizes annotations in JSON files.
}
\label{data-structure}
\end{figure}

\textbf{Directory Tree.}
The dataset is structured into three subsets: training, validation, and
test, each containing images and corresponding annotation files, shown in
Figure \ref{data-structure}.
%
%
The directory follows a \textsc{YOLO}v7-v12 hierarchical format, where images and
their respective labels are stored in separate folders within each subset.
The \frcnn data follow the \textsc{COCO} formatting, where all labels are stored in
a single \texttt{JSON} file for each set.

\begin{figure}[ht]
\centering
\includegraphics[width=1\linewidth]{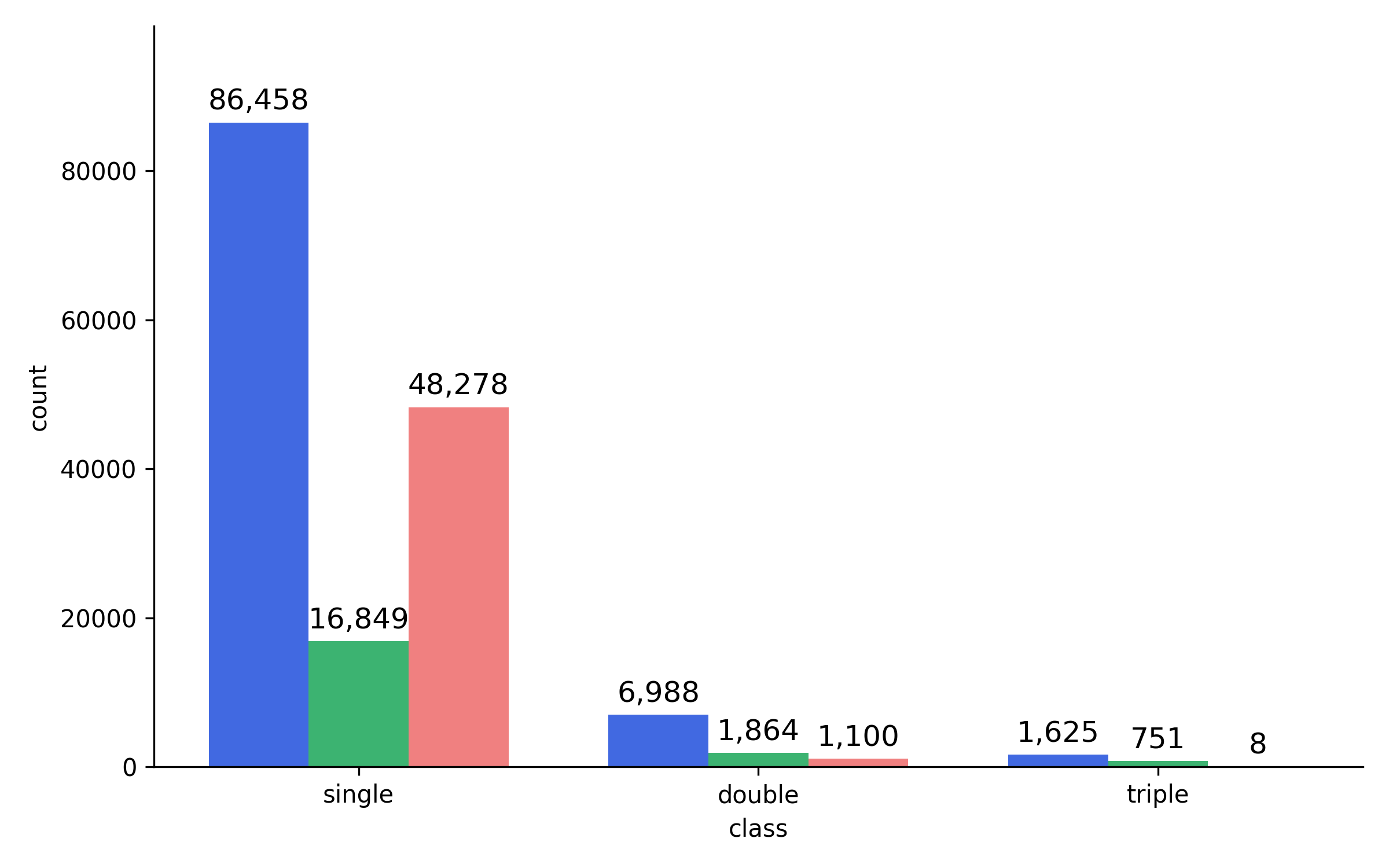}
\caption{Class distribution across training, validation, and test datasets.
The bar chart illustrates the number of instances for each class:
single, double, and triple within the three datasets.
The training set is represented in shades of blue,
the validation set in shades of green,
and the test set in shades of red, with darker shades indicating
single and progressively lighter shades for double and triple. }
\label{bar-dist}
\end{figure}

\textbf{Dataset Distribution.} 
%
%
%
%
%
%
%
%
%
%
%
%
Figure \ref{bar-dist} summarize the distribution of annotated
maize seedlings across the training, validation, and test subsets.
The dataset contains a total of 163,921 annotated objects, with single
plants dominating at 92.47\% of all annotations.
Double plants are notably less frequent at 6.07\%, while triple plants are
rare, comprising only 1.45\%, with the lowest occurrence in the test
set.
This imbalance in plant distribution, particularly the scarcity of
clustered plants, may affect detection performance due to occlusion and
overlap.
The visualization further highlights variations across subsets and
potential dataset biases.

%

\begin{figure*}[!t]
\centering

\subfloat[V2--V4, light-colored soil]{%
    \includegraphics[width=0.27\linewidth]{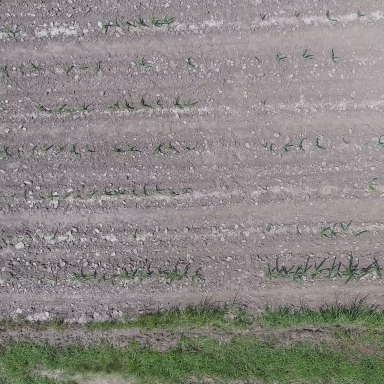}%
    \label{1}
}
\hfil
\subfloat[V4--V6, reddish soil]{%
    \includegraphics[width=0.27\linewidth]{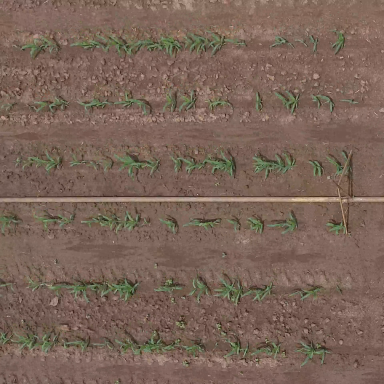}%
    \label{4}
}
\hfil
\subfloat[different genotypes, wet soil]{%
    \includegraphics[width=0.27\linewidth]{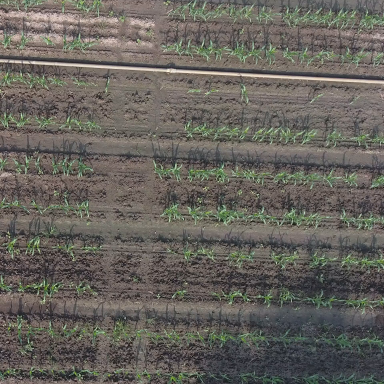}%
    \label{10}
}

\vspace{-0.1mm}

\subfloat[V4, low altitude]{%
    \includegraphics[width=0.27\linewidth]{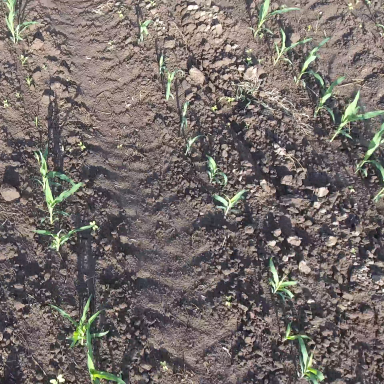}%
    \label{5}
}
\hfil
\subfloat[overcast, V10]{%
    \includegraphics[width=0.27\linewidth]{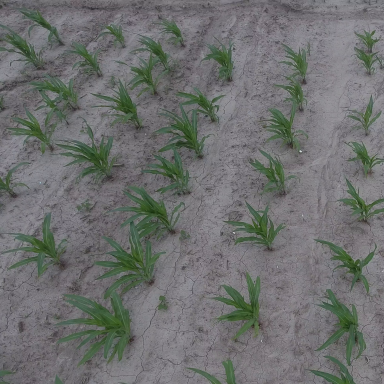}%
    \label{9}
}
\hfil
\subfloat[V12, cluster]{%
    \includegraphics[width=0.27\linewidth]{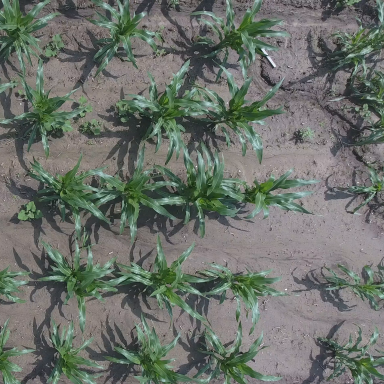}%
    \label{11}
}

\vspace{-0.1mm}

\subfloat[V2--V4, reddish soil]{%
    \includegraphics[width=0.27\linewidth]{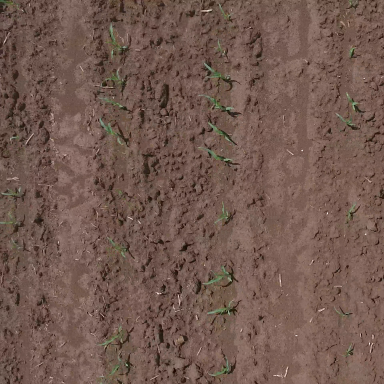}%
    \label{2}
}
\hfil
\subfloat[V4--V6, overcast]{%
    \includegraphics[width=0.27\linewidth]{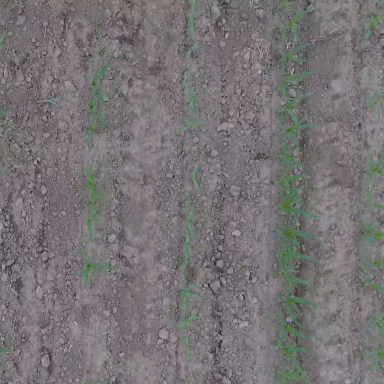}%
    \label{6}
}
\hfil
\subfloat[high altitude]{%
    \includegraphics[width=0.27\linewidth]{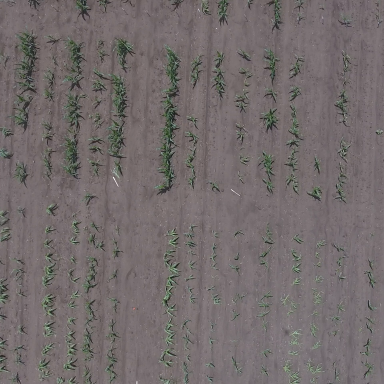}%
    \label{7}
}

\vspace{-0.1mm}

\subfloat[V8--V12, light-colored soil]{%
    \includegraphics[width=0.27\linewidth]{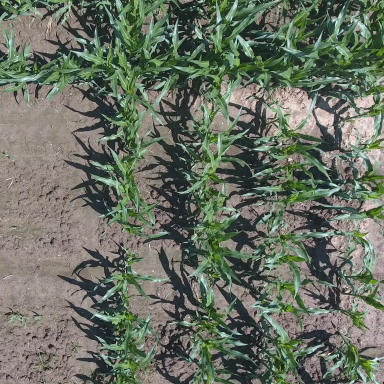}%
    \label{3}
}
\hfil
\subfloat[longer shadows]{%
    \includegraphics[width=0.27\linewidth]{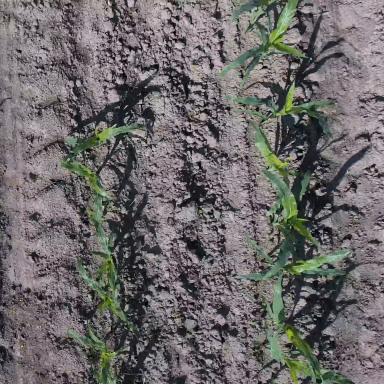}%
    \label{8}
}
\hfil
\subfloat[high altitude, longer shadows]{%
    \includegraphics[width=0.27\linewidth]{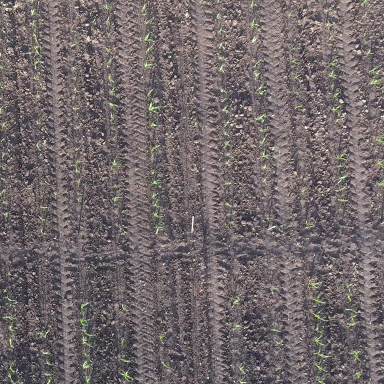}%
    \label{12}
}
\caption{Sample data variability, illustrating differences in maize lines, growth stages, soil colors, planting densities, camera angles, altitudes, and plant orientations. The first row shows UAV views perpendicular to the crop rows; the second row presents oblique camera views; the third row depicts UAVs flying parallel to the crop rows; and the fourth row captures variations across different times of day.}
\label{variability}
\end{figure*}

\textbf{Data Variability.}
Figure \ref{variability} illustrates the environmental variability present
in our training, validation, and test datasets.
Soil color varies due to rainfall, irrigation, time of day, and soil
composition, all of which affect the contrast between seedlings and the
background.
Higher UAV flight altitudes result in smaller, less-defined seedlings due
to reduced pixel resolution.
Lighting conditions range from sunny to cloudy, each
affecting image contrast and seedling visibility differently.
Time of day further influences shadow length and orientation, altering
the visual characteristics of the field.
Wind conditions impact plant morphology, causing leaves to bend or
appear deformed.
Camera angle variations, from nadir (top-down) to oblique views, affect
visibility, occlusion, and the apparent shape of seedlings.
In addition to environmental factors, biological
%
%
%
variation also
challenges detection: early-stage seedlings (V1--V3) can resemble weeds,
while later stages (V8--V12) tend to overlap and occlude.
Genotypic diversity
%
%
among the mutant, inbred, and elite maize lines
introduces structural variation in leaf shape and plant architecture.


\subsection{Labeling Challenges}
\label{sec:orgfa46576}

\begin{figure*}[!t]
\centering

\subfloat[semi-automatic labeling]{%
    \includegraphics[width=0.475\linewidth]{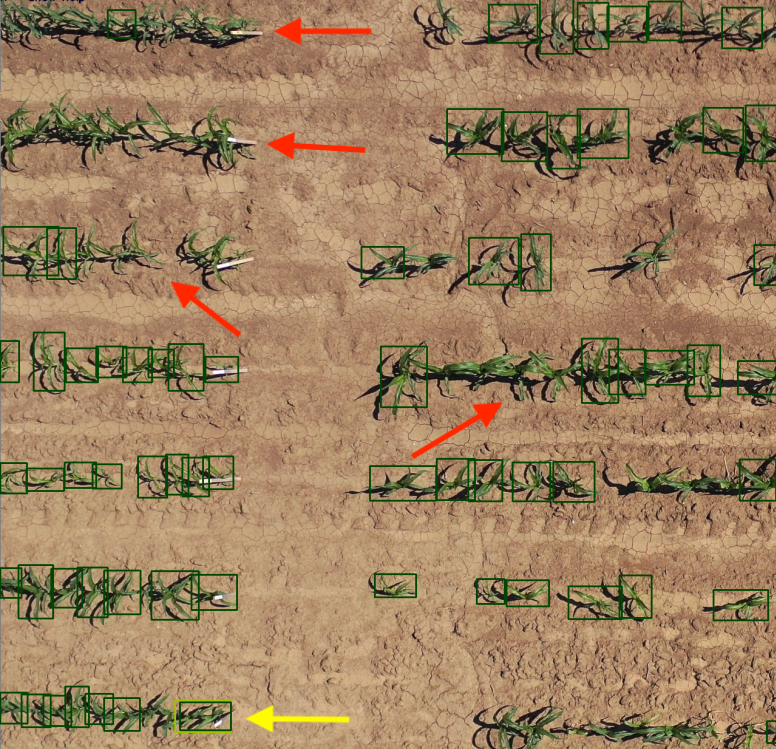}%
    \label{semiauto}
}
\hfil
\subfloat[corrected]{%
    \includegraphics[width=0.475\linewidth]{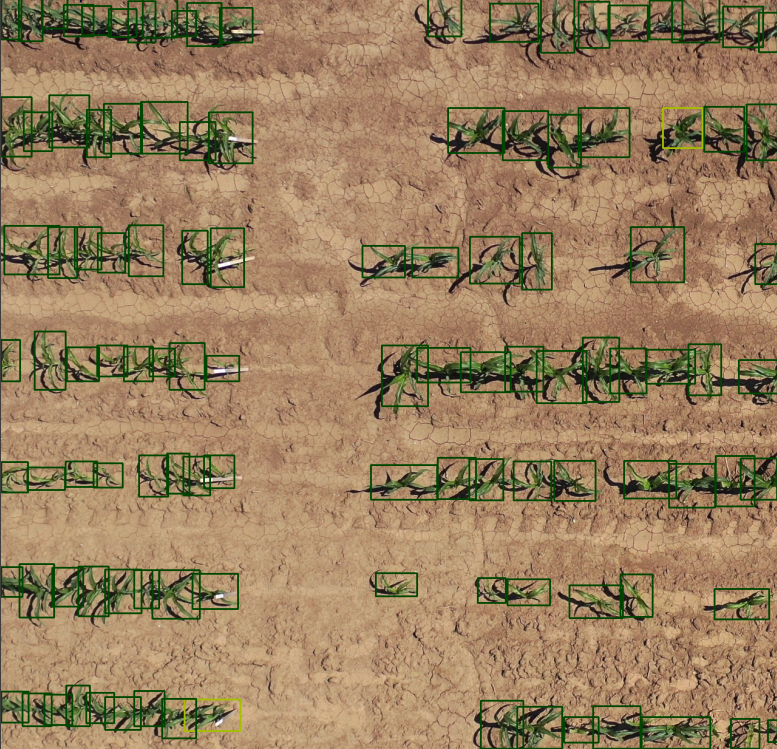}%
    \label{semicorrect}
}

\caption{Semi-automatic annotation using YOLO on maize plants at growth stages V8--V10.
While the model successfully detects most plants, some instances are either missed or incorrectly
labeled. Manual correction was applied to improve labeling accuracy.
Yellow arrows indicate duplicate detections (single and double) for the same plant,
where the incorrect bounding box (single) was removed.
Red arrows highlight missed detections that were manually added during the correction process.
}
\label{f-label}
\end{figure*}

The dataset was iteratively refined to improve annotation quality, minimize
redundancy, and enhance detection performance, establishing a high-quality
benchmark for maize seedling detection.
The annotation process was
conducted in three iterative stages, each progressively improving the
accuracy and robustness of the labeled dataset.
The first stage involved manual labeling of an initial dataset (\(D_m\)) to
provide ground truth annotations (Table \ref{train-data}).
This manually labeled set was then used to train an initial detection
model
%
%
that was leveraged in the second stage to generate additional
labels automatically.
The second stage introduced a \yolo-assisted labeling approach, where a new
set of 3,182 frames was processed, and model predictions were saved as
tentative annotations.
However, several systematic detection failures were observed during this
process.
On average, 30\% of maize plants per frame were not detected, especially at
higher altitudes and during growth stages where seedlings were either too
small (before V3) or too large (beyond V8), leading to weak visual signals
or overlapping foliage that obscured individual stands (see Figure
\ref{f-label}).
The model also frequently misidentified small plant fragments at frame
edges, misclassified shadow artifacts as maize seedlings, and failed to
detect low-resolution plants.
To address these issues, a manual quality control process was implemented
in which the annotator reviewed and corrected bounding boxes frame by
frame.
In the third stage, a homography-based labeling approach was introduced to
further improve annotation efficiency and accuracy.
%
%
%
Because consecutive video frames contain overlapping regions, the same
maize plant often appears in multiple frames, requiring redundant
annotations.
To minimize this, we computed a homography for each successive pair of
frames by extracting feature descriptors with \asift and estimating a
transformation matrix from the matched keypoints.
This matrix was then used to project bounding boxes forward from one frame
to the next, reducing the need for repetitive manual labeling.
Newly appearing maize plants were manually marked, while previously labeled
plants were transferred forward across frames.
To ensure accuracy, annotations were verified in both forward and backward
projection, eliminating inconsistencies in overlapping regions.
This method significantly reduced manual annotation effort while preserving
spatial consistency across sequential frames.
%


%
Another key challenge stemmed from input resolution constraints. The \yolo
model accepts
\(640 \times 640\) images, whereas the original dataset resolution
was \(4096 \times 2160\) and \(3840 \times 2160\).
Downscaling distorts plant shapes due to aspect ratio
compression, reducing annotation precision.
To mitigate this issue, the dataset was fragmented into square image
patches that aligned with the \yolo input resolution.
Initially, fragmenting into \(640 \times 640\) patches resulted in 28,000
patches, but this proved ineffective for low-altitude \uav images, as
seedlings were often split across multiple patches, resulting in incomplete
morphology and poor detections.
To preserve plant structures, an optimized patch size of \(1920 \times 1920\) was
adopted, with \(1280 \times 1280\) patches considered for data augmentation.
Since \yolo accepts inputs of \(640 \times 640\), these larger patches were
downsampled to fit the network's input size, retaining the overall
morphology of the plants but sacrificing some spatial resolution.
%

\begin{figure*}[!t]
\centering
\subfloat[original label]{%
    \includegraphics[width=0.3\linewidth]{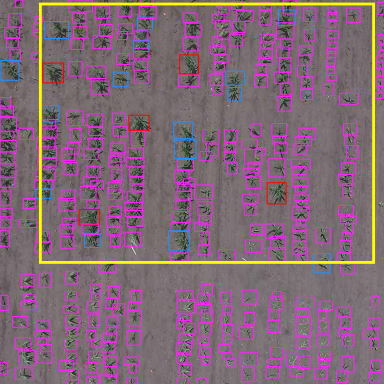}%
    \label{ori-lbl}
}
\hfil
\subfloat[projection with homography]{%
    \includegraphics[width=0.3\linewidth]{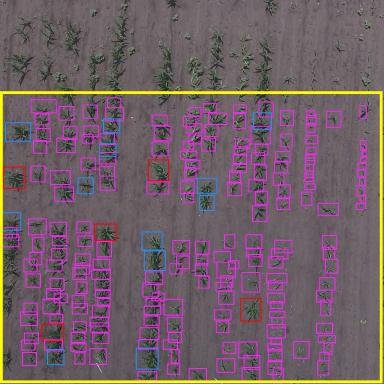}%
    \label{proj-lbl}
}
\hfil
\subfloat[manual correction]{%
    \includegraphics[width=0.3\linewidth]{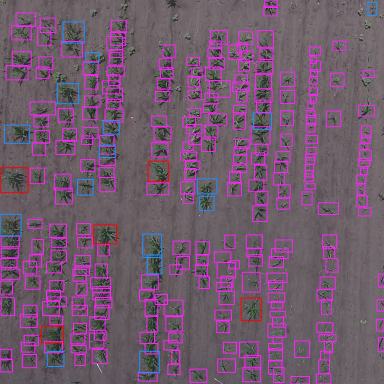}%
    \label{cr-lbl}
}
\caption{
Bounding box projection using homography to accelerate the labeling process.
Existing annotations from Frame A are projected onto Frame B based on spatial overlap.
The yellow box indicates the region of overlap between the two frames.
Only newly visible objects in Frame B are manually labeled.
}
\label{homography_label-proj}
\end{figure*}

%
Figure \ref{homography_label-proj}
provides visual examples, including a manually
labeled image from the first set, a failed detection case from the second
set (highlighting high-altitude and early-stage maize errors), and an
example of manual annotation with homography-based projection before and
after correction.


\subsection{Training with \yv12}
\label{sec:org430e8ab}

\begin{figure*}[!t]
\centering
\includegraphics[width=0.8\textwidth]{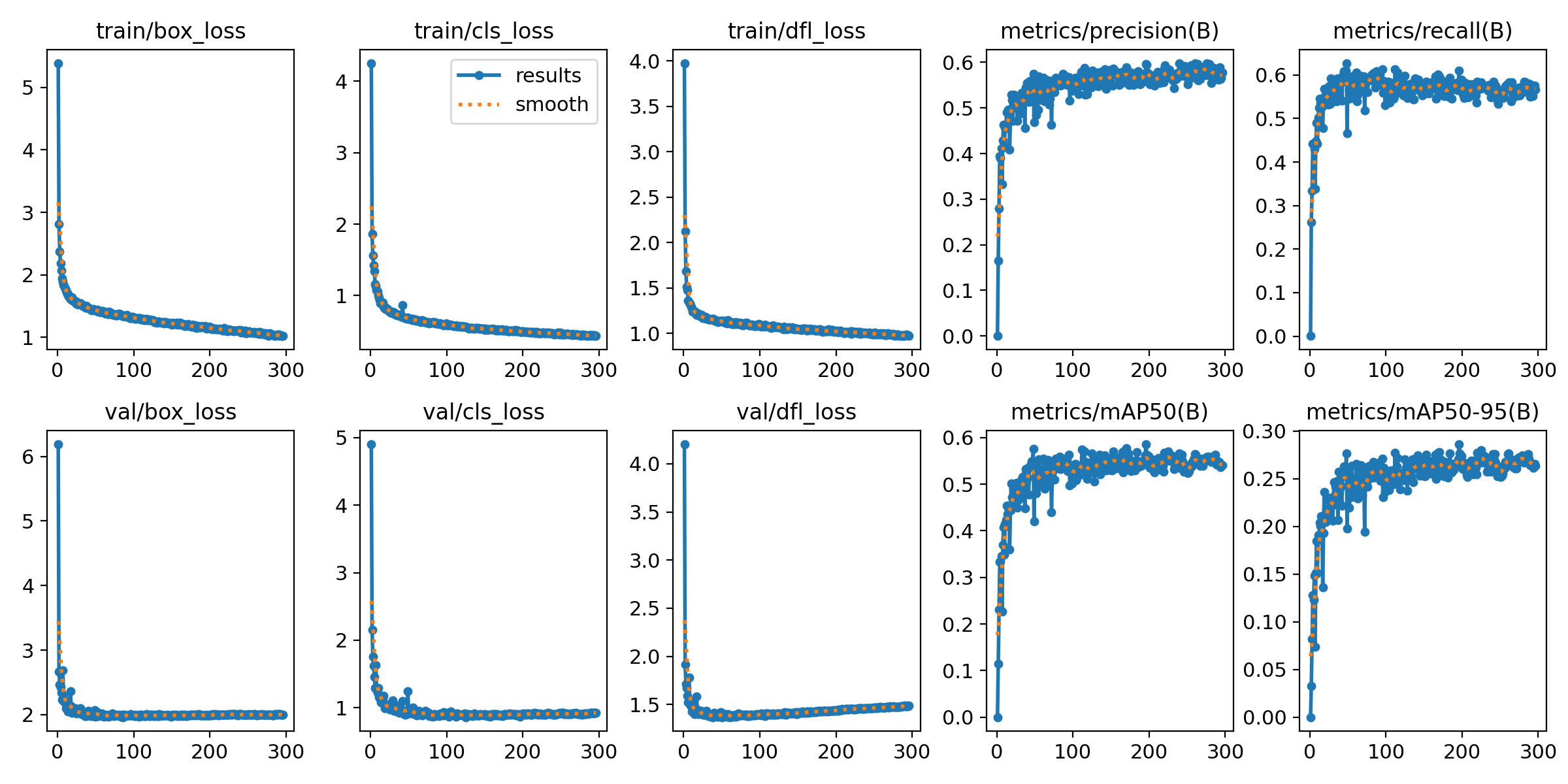}
\caption{Training and validation curves for object detection. 
box\_loss: bounding box regression loss, 
cls\_loss: classification loss, 
dfl\_loss: distribution focal loss (for improved localization), 
precision(B): proportion of correct detections, 
recall(B): proportion of ground truth objects detected, 
mAP50(B): mean average precision at IoU $\geq$ 0.50, 
mAP50-95(B): mAP averaged over IoU thresholds from 0.50 to 0.95. 
All metrics are computed on the validation set. 
The decreasing trend of loss indicates model optimization, while the increasing metrics indicate effective learning and performance improvement.
}
\label{train-loss-met}
\end{figure*}

%
We report training metrics from the various \yv12 models.
%
%
The \yv12 model was trained following the methodology detailed in Section
\ref{train-method}.
Throughout training, training loss and validation
performance were monitored to assess model convergence and generalization.
Figure \ref{train-loss-met}
presents several metrics
over epochs both for
training (top row) and validation (bottom row).
The training and validation curves indicate that the model is effectively
learning and converging.
The decreasing loss values for bounding box regression, classification, and
distribution focal loss confirm that the model is optimizing its
predictions over time.
The increasing precision, recall, and mean Average Precision (\map) metrics further suggest that the
model's ability to detect and classify objects is improving with each
epoch.
%
%
However, the \map for validation remains relatively
low, which indicates that while the model is learning, it struggles with
generalizing well to unseen data.
This could be due to factors such as class imbalance or the need for
further hyperparameter tuning to enhance detection accuracy.

The low \map indicates that although the model is detecting objects, the
overall quality of its predictions, particularly in terms of localization
accuracy and classification confidence, may require further improvement.
The validation loss shows a slight increase than the training loss,
indicating a small generalization gap.
This suggests that while the model does not suffer from severe overfitting,
there is still room for improvement in its ability to generalize beyond the
training dataset.
Fluctuations in validation metrics are expected, but the
overall trend remains stable and shows progressive improvement.
%
%
To enhance the model's validation performance, further tuning may be
required, such as adjusting anchor sizes, using data augmentation
techniques, or incorporating more diverse training samples.
While the current results indicate effective learning, additional
refinements could be necessary to boost validation \map and ensure the model
performs well in real-world scenarios.

\begin{figure*}[!t]
\centering
\subfloat[\yoloe]{%
    \includegraphics[width=0.475\linewidth]{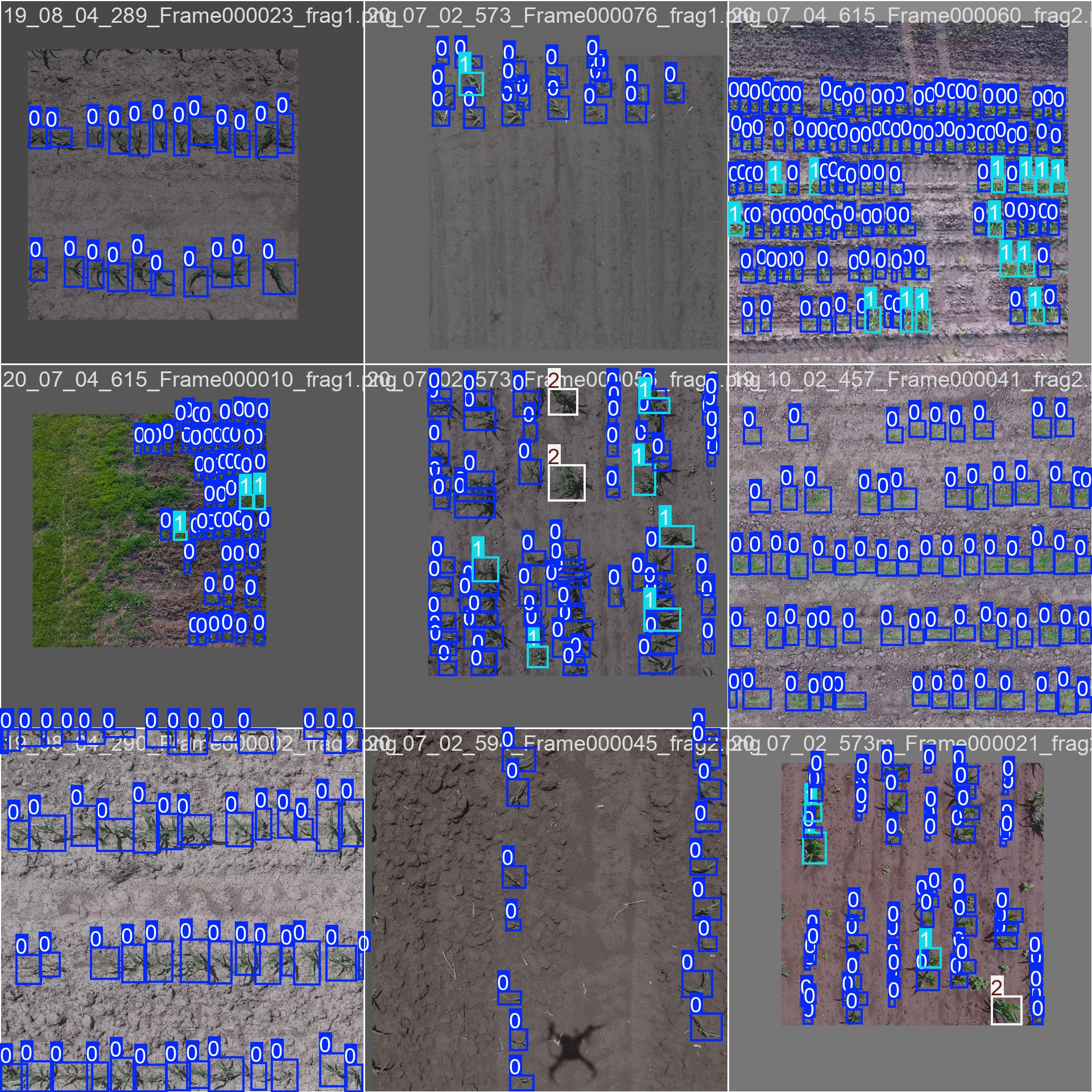}%
    \label{b0}
}
\hfil
\subfloat[\yolot]{%
    \includegraphics[width=0.475\linewidth]{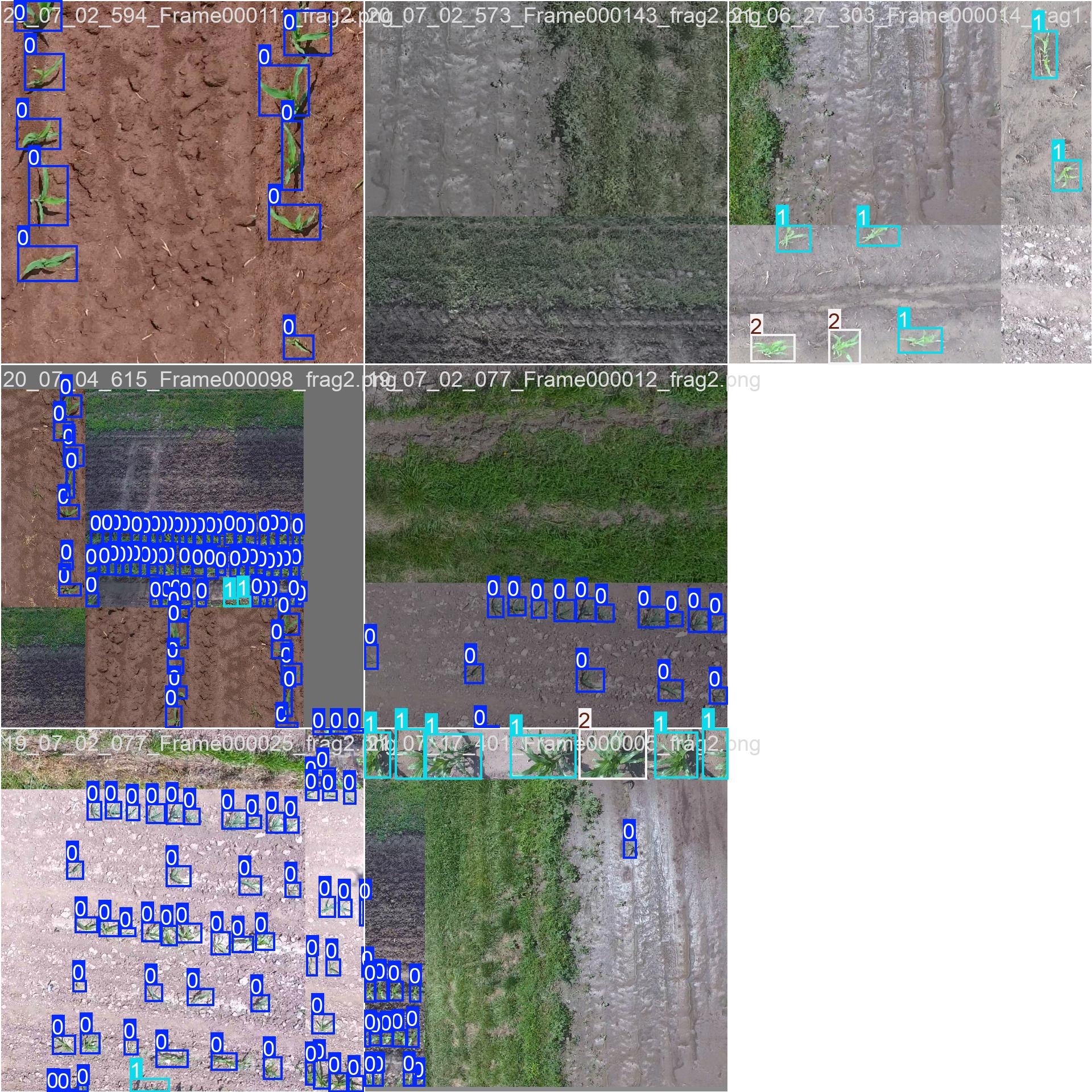}%
    \label{b1}
}
\caption{Sample image batches used during training.
The maximum batch size that our Lambda Lab machine can handle varies by model.
\yoloe can process 18 images per batch, utilizing 9 images per GPU,
while \yolot can process 14 images per batch, with 7 images per GPU.
During training, mosaic augmentation is disabled for \yoloe but enabled for \yolot.
Mosaic augmentation combines fragments from multiple images into a single training sample,
effectively increasing dataset diversity, improving generalization, and helping the model
learn to detect objects under varied contexts and scales.
}
\label{tr-batch}
\end{figure*}

\subsection{Detection Performance and Accuracy Metrics}
\label{sec:orgdab7338}

\begin{figure*}[!t]
\centering
\subfloat[\yolon]{%
    \includegraphics[width=0.46\linewidth]{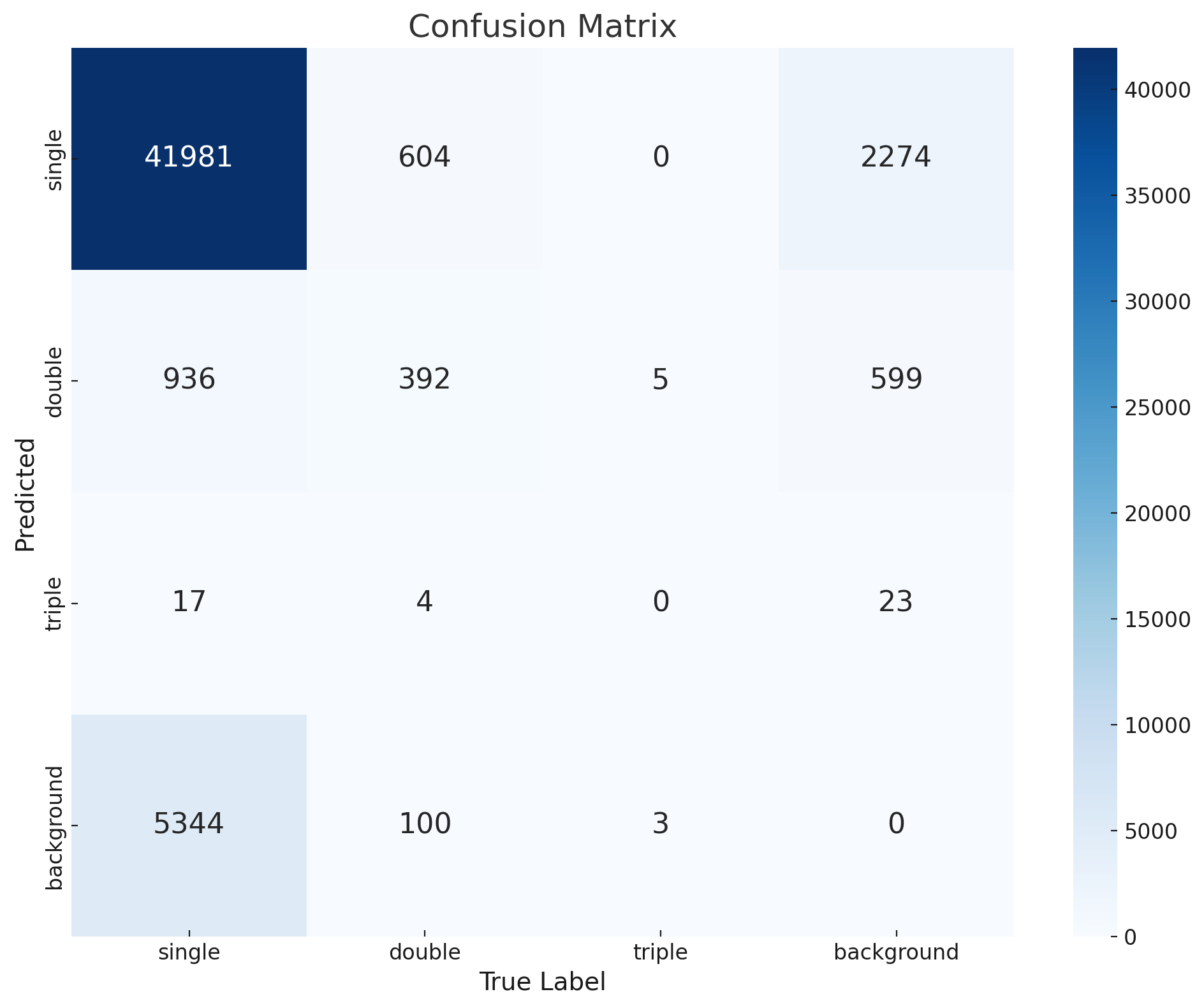}%
    \label{confuse-n}
}
\hfil
\subfloat[\yoloe]{%
    \includegraphics[width=0.46\linewidth]{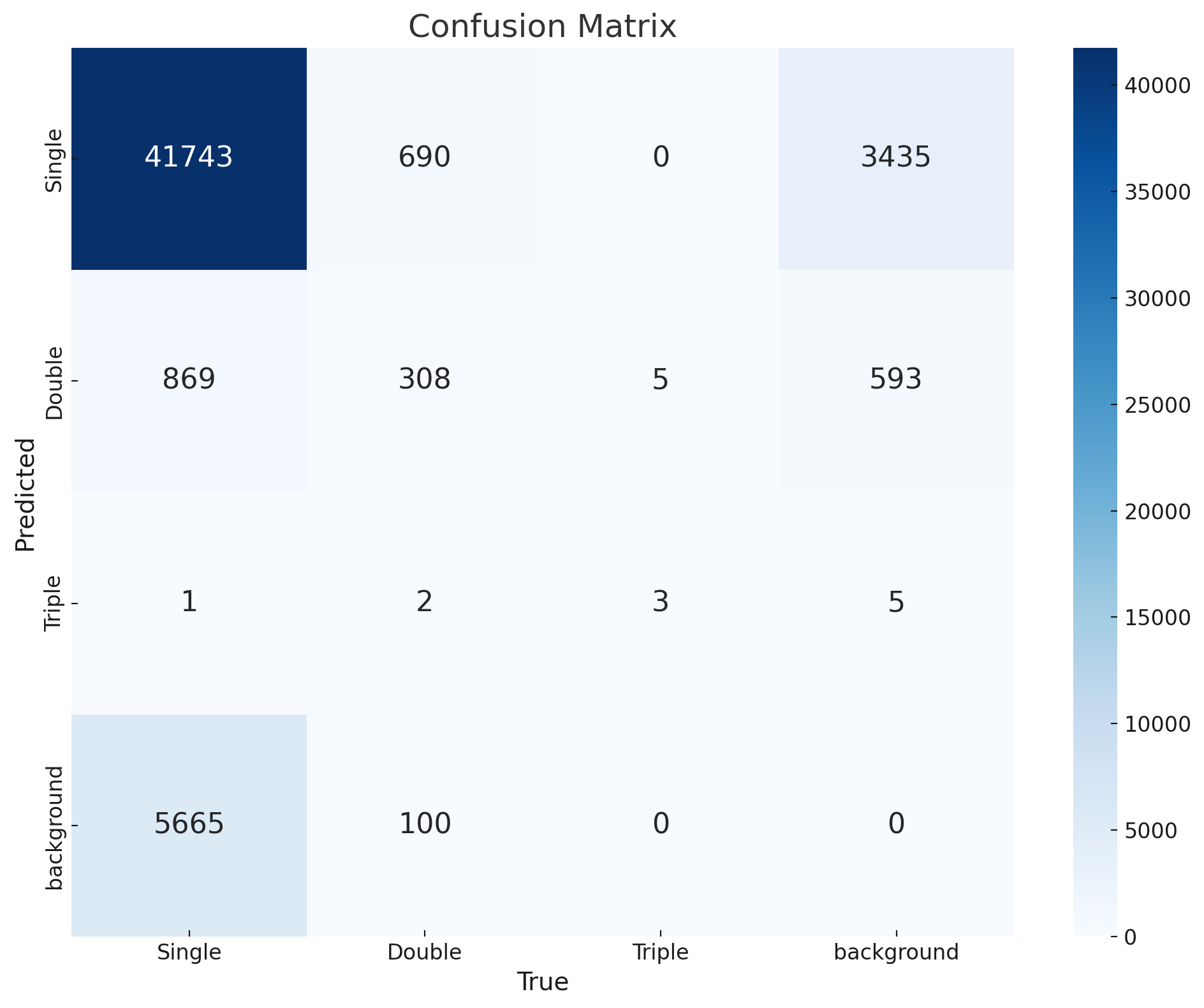}%
    \label{confuse-e}
}

\vspace{1mm}

\subfloat[\yolot]{%
    \includegraphics[width=0.46\linewidth]{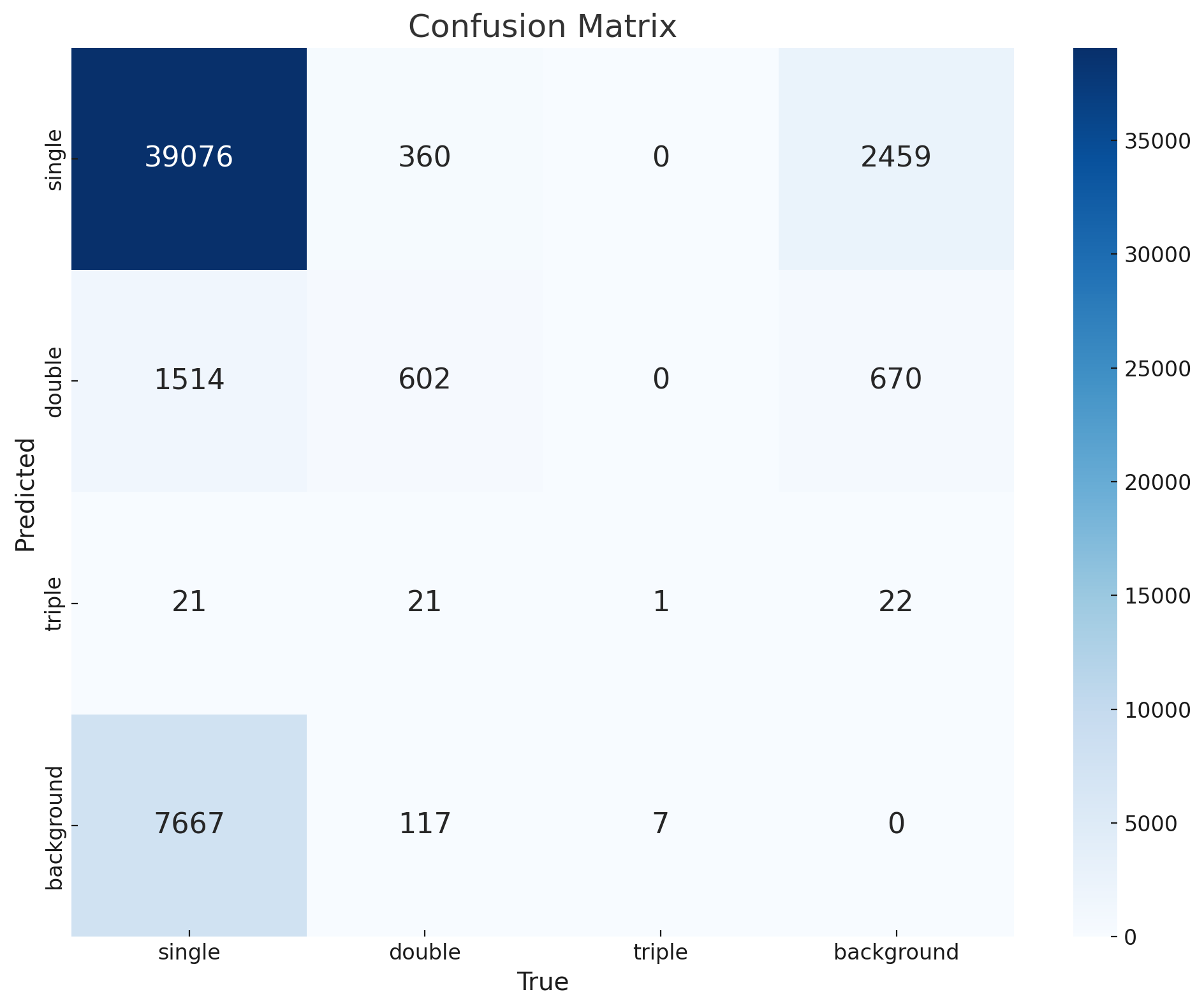}%
    \label{confuse-t}
}
\hfil
\subfloat[\frcnn]{%
    \includegraphics[width=0.46\linewidth]{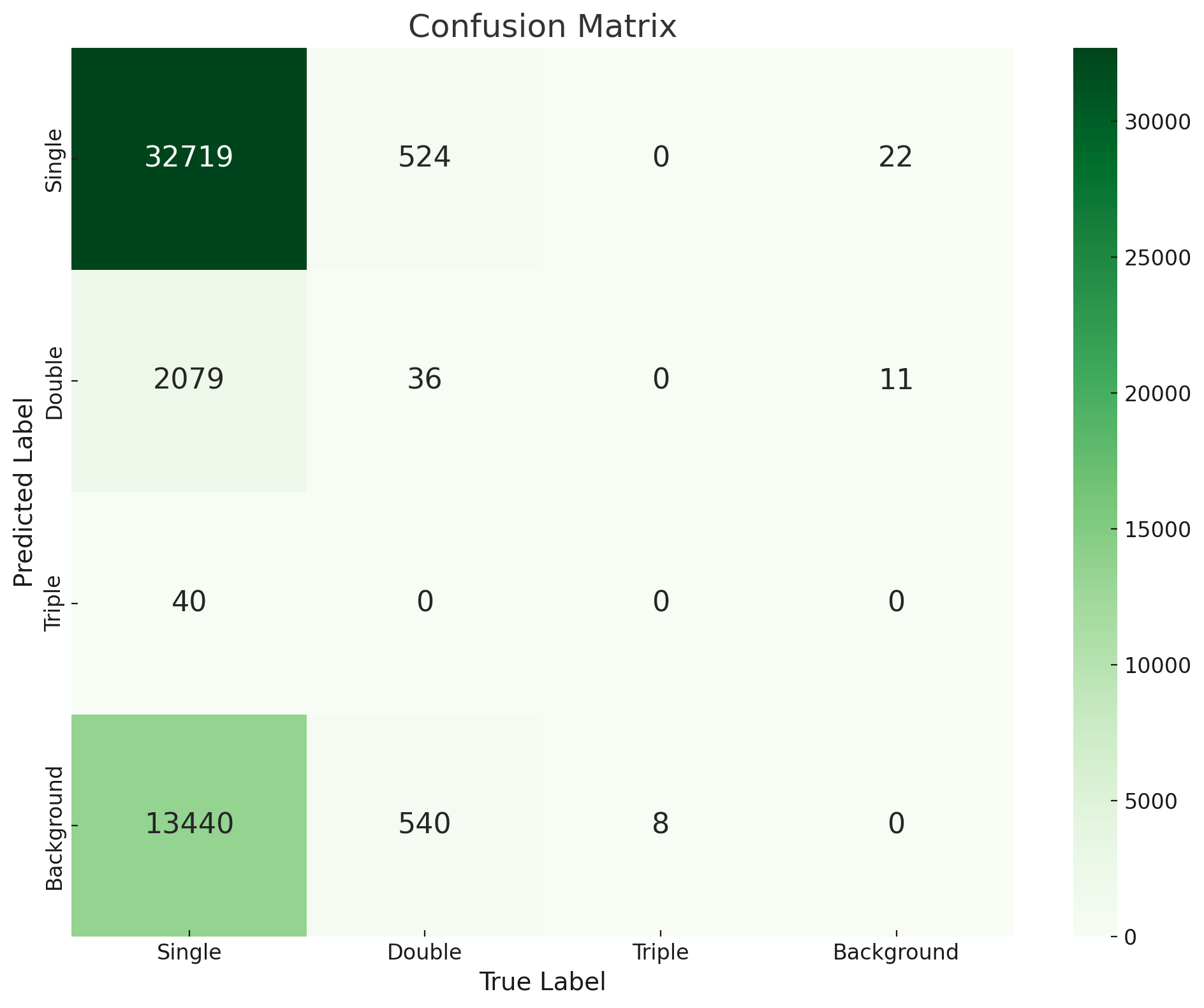}%
    \label{confuse-r}
}
\caption{Confusion matrices (non-normalized) for each model.
Each matrix shows the count of true positives, false positives, false negatives,
and true negatives across the single, double, and triple object classes.
The diagonal elements indicate correctly classified instances,
while off-diagonal values reveal misclassifications.
The results highlight the models' varying performance,
particularly on under-represented double and triple classes.
}
\label{confuse}
\end{figure*}

Table \ref{matrices} and \ref{matrices-2}
presents the quantitative evaluation of four object
detection models: \yolon, \yoloe, \yolot, and \frcnn. The evaluation
includes
%
%
\map
%
at IoU threshold 0.50 (mAP@50) and averaged
over IoU thresholds from 0.50 to 0.95 (mAP@50:95), along with precision,
recall, and F1.
These metrics are reported separately across three object grouping
categories: single, double, and triple, reflecting increasing complexity in
detection scenarios.
Across all models, detection performance is significantly higher in the
single-class category.
This is
%
%
expected
given the significant class imbalance in
the dataset.
The overwhelming presence of single-object samples biases the
models toward the dominant class, limiting their ability to generalize and
perform accurately on the less represented double and triple classes.
This limitation is further illustrated in the confusion matrix shown in
Figure \ref{confuse},
%
%
which presents the true positives, true negatives,
false positives, and false negatives for each class and highlights the
discrepancies in classification performance.
All four models exhibit a recurring pattern of misclassification involving
the single plant and background.
A substantial number of true single
plants were misclassified as background, indicating the rate of missed
detections.
Conversely, background regions were sometimes misclassified as
single plants, resulting in a considerable number of false positives.
Among the models, the \frcnn model showed the highest rate of missed
detections, with over 13,000 single plants incorrectly predicted as
background.
In contrast, the \yolo
models demonstrated improved performance
in this regard, with \yolon reducing this error to approximately 5,300
instances.
These misclassifications primarily occurred in cases where the
plants were either too small or excessively large, making them harder to
detect accurately.
False positives often occurred in fragmented plant regions that were not
labeled as plants because only a small portion was visible.
This suggests that object scale and annotation granularity play a critical
role in the observed classification errors.

Across the evaluated models, clear trade-offs emerged between precision,
recall, and overall detection robustness.
\frcnn demonstrated the highest precision on single class detection, but
this came at the cost of substantially lower recall, which limited its
overall effectiveness and generalization to more complex multiple plant
clusters.
In contrast, \yolon struck a better balance, combining strong precision
with the highest recall, resulting in the best F1 performance and the
strongest mAP@0.5 for single plants.
However, its performance degraded sharply in more complex double and triple
plant classes, where detection scores dropped to near zero.
\yoloe, while slightly less accurate for single plants, proved more
resilient under increased scene complexity, outperforming other models in
triple class detection.
\yolot showed the most balanced behavior across single and double classes,
achieving the best recall and mAP@0.5 for doubles, suggesting greater
robustness under moderate class imbalance.
Taken together, the results highlight a spectrum: \frcnn excels in
precision but struggles with recall and complexity; \yolon offers the
strongest single class performance but poor generalization; \yoloe adapts
better to challenging scenes; and \yolot provides the most consistent
balance across conditions.

In terms of computational efficiency, Table \ref{exec-time}
presents the
total execution time required for both evaluation and detection across 806
test images.
Among the models, \yoloe was the fastest, completing detection
in 125 seconds and evaluation in 29 seconds.
\yolot followed with detection and evaluation times of 165 and 33 seconds
respectively.
\yolon, while achieving the best detection accuracy for single-object
instances, required 243 seconds for detection and 61 seconds for
evaluation.
\frcnn was the slowest, with 684 seconds for detection and 105 seconds for
evaluation, reflecting the computational demands of its two-stage
architecture.
The detection time includes not only model
inference but also the process of saving prediction images and labels.
These label files are intended for future processing or
downstream analysis, which contributes to the longer detection time
compared to evaluation alone.
As expected, the \yolo-based models processed the test data significantly
faster than \frcnn, reinforcing their suitability for real time or
near-real time applications.

\begin{figure*}[!t]
\centering
\subfloat[good brightness]{%
    \includegraphics[width=0.3\linewidth]{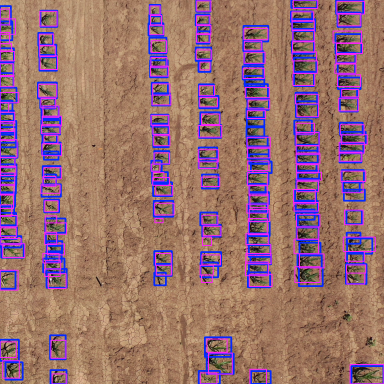}%
    \label{gb}
}
\hfil
\subfloat[double trouble]{%
    \includegraphics[width=0.3\linewidth]{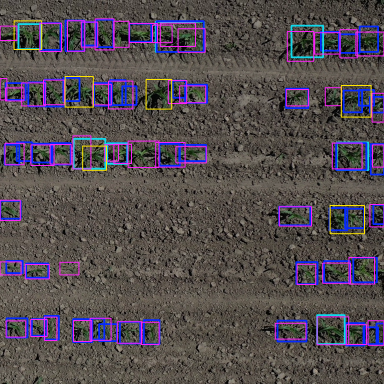}%
    \label{doubletrb}
}
\hfil
\subfloat[overcast]{%
    \includegraphics[width=0.3\linewidth]{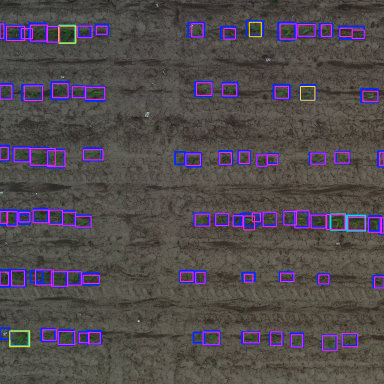}%
    \label{uc}
}

\vspace{1mm}

\subfloat[plant \emph{vs} weed]{%
    \includegraphics[width=0.3\linewidth]{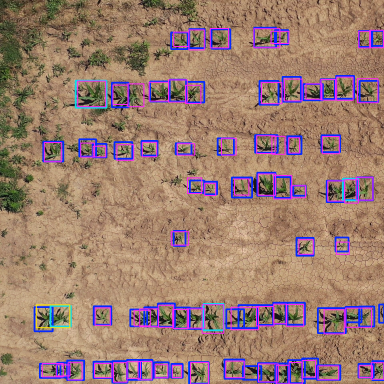}%
    \label{pweed}
}
\hfil
\subfloat[small plants]{%
    \includegraphics[width=0.3\linewidth]{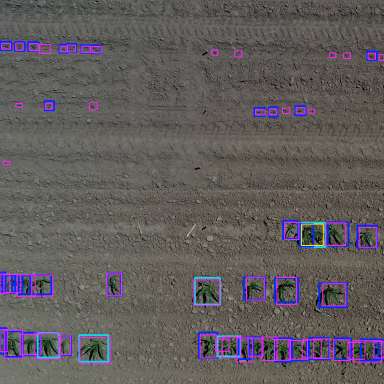}%
    \label{small}
}
\hfil
\subfloat[multiple detections]{%
    \includegraphics[width=0.3\linewidth]{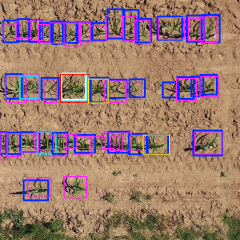}%
    \label{mdetect}
}

\vspace{1mm}

\subfloat[soil variation]{%
    \includegraphics[width=0.3\linewidth]{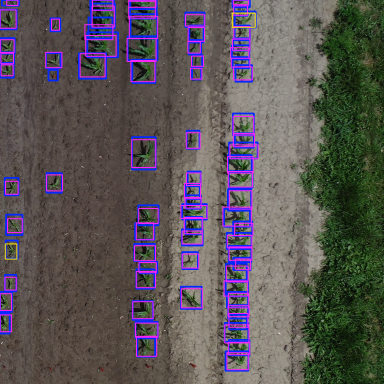}%
    \label{soil}
}
\hfil
\subfloat[oblique small plants]{%
    \includegraphics[width=0.3\linewidth]{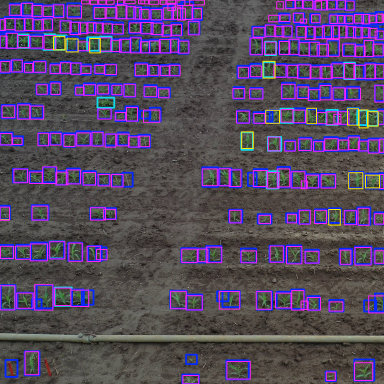}%
    \label{obs}
}
\hfil
\subfloat[oblique big plants]{%
    \includegraphics[width=0.3\linewidth]{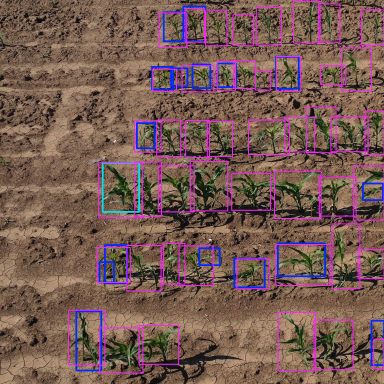}%
    \label{obb}
}
\caption{Sample comparison of \yoloe predictions and ground truth,
illustrating challenges related to growth stages, camera angles,
flight altitudes, multiple detections, soil colors, and lighting conditions.
\yoloe predictions: single (dark blue), double (light blue), triple (lighter blue).
Ground truth: single (magenta), double (yellow), triple (red).
}
\label{result-problems}
\end{figure*}

\begin{table*}
\caption{
The first evaluation metrics (mAP@0.5 and mAP@0.5:0.95) for \yolon, \yoloe, \yolot, and \frcnn across single, double, and triple classes. 
mAP@0.5 refers to the mean Average Precision at an IoU threshold of 0.5, indicating how well the predicted bounding boxes overlap with the ground truth. 
mAP@0.5:0.95 averages the precision over IoU thresholds between 0.5 and 0.95, providing a more comprehensive measure of localization and classification performance. 
The best performance in each category is shown in bold.
}
\centering
\begin{tabular}{|l|c|c|c|c|c|c|}
\hline
\textbf{Model}  & \multicolumn{3}{c|}{\textbf{mAP@0.5}} & \multicolumn{3}{c|}{\textbf{mAP@0.5:0.95}} \\ 
\hline
              & Single          & Double           & Triple          & Single          & Double          & Triple \\ 
\hline
\yolon        &   \textbf{0.916} &  0.233           &          0.000  &  \textbf{0.568} & \textbf{0.138}  &          0.000  \\ 
\yoloe        &   0.901          &  0.191           &  \textbf{0.313} &          0.507  &         0.110   &  \textbf{0.156} \\ 
\yolot        &   0.887          &  \textbf{0.253}  &          0.025  &          0.542  &         0.137   &          0.012  \\ 
\frcnn        &   0.655          &  0.106           &          0.000  &          0.299  &         0.055   &          0.000  \\ 
\hline
\end{tabular}
\label{matrices}
\end{table*}

\begin{table*}
\caption{The second evaluation metrics (precision, recall, and F1) for \yolon, \yoloe, \yolot, and \frcnn across single,
double, and triple classes.
Precision measures the proportion of correct predictions among all predicted instances.
Recall measures the proportion of correct predictions among all ground truth instances.
and F1  is the harmonic mean of Precision and Recall, balancing both aspects. 
The best performance is shown in bold.}
\centering
\begin{tabular}{|l|c|c|c|c|c|c|c|c|c|}
\hline
\textbf{Model}  & \multicolumn{3}{c|}{\textbf{Precision}} & \multicolumn{3}{c|}{\textbf{Recall}} & \multicolumn{3}{c|}{\textbf{F1}} \\
\hline
                & Single & Double & Triple & Single & Double & Triple & Single & Double & Triple \\
\hline
\yolon          &          0.940  &          0.243  &         0.000  & \textbf{0.873} &         0.427  &         0.000  & \textbf{0.905} & 0.310          & 0.000           \\
\yoloe          &          0.912  &          0.236  & \textbf{0.397} &         0.855  &         0.349  & \textbf{0.500} & 0.883          & 0.282          & \textbf{0.443}  \\
\yolot          &          0.953  &  \textbf{0.285} &         0.030  &         0.774  & \textbf{0.528} &         0.125  & 0.854          & \textbf{0.370} & 0.048           \\
\frcnn          &  \textbf{0.984} &          0.017  &         0.000  &         0.678  &         0.033  &         0.000  & 0.803          & 0.022          & 0.000           \\ 
\hline
\end{tabular}
\label{matrices-2}
\end{table*}

\begin{table*}[ht]
\centering
\caption{Execution time (in seconds) to run evaluation and detection on 806 images from the test dataset. The best performance is shown in bold.}
\begin{tabular}{|l|r|r|r|}
\hline
\textbf{Model} & \textbf{Evaluation (s)} & \textbf{Evaluation fps} & \textbf{Detection (s)} \\
\hline
\yolon  &          61  &         13  &         243  \\
\yoloe  &  \textbf{29} & \textbf{27} & \textbf{125} \\
\yolot  &          33  &         24  &         165  \\
\frcnn  &          105 &          7  &         684  \\
\hline
\end{tabular}
\label{exec-time}
\end{table*}

\subsection{Generalizability Across Different Cases}
\label{sec:org8e17c8d}

%
%
We evaluated which growth stages are most reliably detected and how well
the models generalize to oblique views, despite being trained with limited
oblique data.
Table \ref{general} and \ref{general2}
presents a comparison of \yoloe and \yolot models
evaluated on groups of test cases, including growth
stages V2--V4, V4--V8, V6--V12, and oblique camera views.
Overall, both models perform best on the V4--V8 growth stage across all
metrics and levels of variability.
For example, \yolot achieves the highest
mAP@0.5 of 0.933 and a recall of 1.000 under single and triple conditions,
respectively.
This suggests that the V4--V8 stage is the optimal growth stage for
detection, as it provides distinctive visual features that are easier for
the models to recognize and generalize the plant instances. 
In contrast, the V6--V12 stage consistently yields the lowest performance,
particularly under the triple-plant condition.
For instance, \yolot's mAP@0.5:0.95 drops to 0.000 and Recall to 0.000 in
this case, indicating poor generalization to more mature growth stages
where plants are clustered and touching each other.
%

%
%
%
Both models demonstrate reasonable generalization to 
oblique camera views despite limited exposure during training.
\yoloe achieves a mAP@0.5 of 0.865 and a Recall of 0.762 under the single
condition, while \yolot performs similarly, with slightly higher precision
but lower recall.
This shows that both models can handle oblique views to some extent,
although their performance still lags behind that of nadir views.

\begin{table*}
\caption{
Generalization performance of \yoloe (top) and \yolot (bottom) across different test conditions,
including growth stages (V2--V4, V4--V8, V6--V12) and oblique camera views.
mAP@0.5 refers to the mean Average Precision at an IoU threshold of 0.5. 
mAP@0.5:0.95 averages the precision over IoU thresholds between 0.5 and 0.95.
The best performance is shown in bold.}
\centering
\begin{tabular}{|l|l|c|c|c|c|c|c|}
\hline
\textbf{Model}  &\textbf{Growth Stages}  & \multicolumn{3}{c|}{\textbf{mAP@0.5}} & \multicolumn{3}{c|}{\textbf{mAP@0.5:0.95}} \\ 
\hline
                &       & Single & Double & Triple & Single & Double & Triple \\
\hline
\yoloe & V2--V4        &          0.907  &          0.160  &             -   &          0.482  &  0.11           &            -     \\ 
       & V4--V8        & \textbf{0.928}  & \textbf{0.213}  & \textbf{0.995}  & \textbf{0.559}  &  \textbf{0.128} &  \textbf{0.697}  \\ 
       & V6--V12       &          0.868  &          0.168  &          0.379  &          0.452  &  0.089          &           0.160  \\ 
       & oblique       &          0.865  &          0.209  &             -   &          0.509  &  0.123          &            -     \\ 
\hline
\yolot & V2--V4        &          0.915  & \textbf{0.295}  &             -   &          0.561  &  \textbf{0.166} &            -     \\ 
       & V4--V8        & \textbf{0.933}  &          0.238  & \textbf{0.249}  & \textbf{0.599}  &           0.139 &  \textbf{0.124}  \\ 
       & V6--V12       &          0.817  &          0.292  &          0.000  &          0.454  &           0.141 &           0.000  \\ 
       & oblique       &          0.836  &          0.243  &             -   &          0.517  &           0.128 &            -     \\ 
\hline
\end{tabular}
\label{general}
\end{table*}

\begin{table*}
\caption{
Generalization performance of \yoloe (top) and \yolot (bottom) across various test conditions, including different growth stages (V2--V4, V4--V8, V6--V12) and oblique camera views. Evaluation metrics include Precision, Recall, and F1-Score, each reported for single, double, and triple class predictions. 
Precision measures the proportion of correct predictions among all predicted instances.
Recall measures the proportion of correct predictions among all ground truth instances.
and F1  is the harmonic mean of Precision and Recall, balancing both aspects. 
The best performance values in each category are shown in bold.
}
\centering
\begin{tabular}{|l|l|c|c|c|c|c|c|c|c|c|}
\hline
\textbf{Model}  &\textbf{Growth Stages}  & \multicolumn{3}{c|}{\textbf{Precision}} & \multicolumn{3}{c|}{\textbf{Recall}} & \multicolumn{3}{c|}{\textbf{F1}} \\
\hline
                &       & Single & Double & Triple & Single & Double & Triple  & Single & Double & Triple \\
\hline
\yoloe & V2--V4        &          0.931  &          0.227  &            -    & \textbf{0.860}  &         0.139  &    -            & \textbf{0.894} &         0.172  &           -    \\
       & V4--V8        & \textbf{0.948}  & \textbf{0.326}  &  \textbf{0.880} &          0.841  &         0.229  &  \textbf{1.000} &         0.891  &         0.269  & \textbf{0.936} \\
       & V6--V12       &          0.886  &          0.218  &           0.528 &          0.801  & \textbf{0.330} &          0.429  &         0.841  &         0.263  &         0.473  \\
       & oblique       &          0.909  &          0.263  &            -    &          0.762  &          0.324 &             -   &         0.829  & \textbf{0.290} &              - \\
\hline
\yolot & V2--V4        &          0.958  & \textbf{0.341}  &            -    & \textbf{0.854}  &          0.417  &             -   & \textbf{0.903} &         0.375  &             -  \\
       & V4--V8        & \textbf{0.976}  &          0.316  &  \textbf{0.130} &          0.772  &          0.396  &  \textbf{1.000} &          0.862 &         0.352  & \textbf{0.230} \\
       & V6--V12       &          0.918  &          0.304  &           0.000 &          0.698  & \textbf{0.635}  &          0.000  &          0.793 & \textbf{0.411} &         0.000  \\
       & oblique       &          0.963  &          0.245  &            -    &          0.644  &          0.405  &             -   &          0.772 &         0.305  &             -  \\
\hline
\end{tabular}
\label{general2}
\end{table*}

\section{Discussion}
\label{sec:org02623bb}

The results of this study highlight both the capabilities and limitations
of current object detection models applied to maize seedling datasets under
realistic field conditions.
All models evaluated in this work, including \yolon, \yoloe, \yolot, and
\frcnn, consistently performed well when detecting single maize seedlings,
particularly during the V4 to V6 growth stages and imaged in nadir view.
However, their performance declined significantly when handling
more complex seedling clusters, such as double and triple seedlings.
This drop in accuracy is clearly reflected in the precision, recall, and
mean Average Precision (\map) metrics.
The \smd dataset was intentionally constructed to be diverse
in terms of maize lines, growth stages, vegetation coverage, soil colors,
lighting conditions such as overcast and glare, varying altitudes and
plant resolutions, wind, camera angles, and different \uavs.
This diversity was critical for training models that generalize well across
a wide range of field scenarios, but it also introduced new sources of
variation that challenged model consistency.
The four models are capable of reliably detecting maize seedlings across a
range of conditions, including the ability to distinguish maize from
visually similar weeds (see Panel \ref{pweed}).
%
Results by growth stage suggest that the V4 to V8 window is the most
favorable for automated detection.
During this stage, plants are large enough to exhibit clear structure, leaf
separation, and spacing, but not yet mature enough to begin touching or
occluding each other.
This stage typically lasts between 7 and 10 days under ideal growing
conditions, making it a practical and efficient target period for
\uav-based data collection.
Early-stage detection, such as at V2 to V4, still yields good evaluation
metrics, but due to weak morphological features and similarity to weeds, it
can result in missed detections, lower confidence, or misclassification
(see Panel \ref{small}).
%
Beyond V8, detection becomes increasingly difficult due to overlapping
leaves, plant clustering, and complex shadows.
In these later stages, larger plants often obscure smaller ones, and the
visual distinction between a double seedling at V6 and a single mature
plant at V12 becomes increasingly subtle.

The strong results for single seedlings are largely due to the dataset
distribution.
More than 92.47\% of annotated objects in the dataset are single plants.
In contrast, double and triple seedlings make up only a small portion, with
6.07\% and 1.45\% respectively.
This class imbalance biases the models toward more frequent categories,
resulting in overly optimistic aggregate evaluation metrics.
As a result, even though the overall evaluation numbers appear strong, they
may not provide a fair representation of the model's performance on the
less common double and triple cases, which has led to poorer generalization
and lower detection accuracy.
Another issue influencing evaluation is inconsistency in ground truth
labeling.
When two maize plants are touching or closely overlapping, annotators
sometimes marked them as a double, and other times as two separate single
plants.
As a result, the model occasionally predicted them as one double, two
singles, or both (see Panel \ref{doubletrb}).
%
If both were detected, it was due to non-maximum suppression only
suppressing multiple boxes from the same class, but not across different
classes.
A similar issue occurs with triples, where the model sometimes predicts
them as a single and a double, or as both double and triple (see Panel
\ref{mdetect}).
%
While predicting either two singles or one double
%
%
is acceptable for
stand counting, it penalizes model performance under standard evaluation.
Metrics like \map depend heavily on exact matches between predictions and
ground truth labels.
In real-world planting, the goal is to grow maize as single plants with
consistent spacing.
Double and triple groupings are anomalies that occur
due to dropping more than one seed into a single hole or planting seeds too
close to one another.
Therefore, while it is important to detect and account for these anomalies
during stand counting, their lower representation in the dataset is
realistic.
Still, their scarcity poses a challenge for model learning.
To address this, we intentionally planted doubles and triples in a range to
increase the number of rare cases.
However, the dataset remained unbalanced.
We also generated synthetic samples to help improve detection accuracy.
The synthetic samples were created by copying masked-out plants and placing
them beside others to simulate double or triple instances, but the
resulting images often lacked a natural appearance due to simplistic
compositing.
Future work could explore generative methods such as GANs or
domain-adaptive simulation to improve realism in rare-class augmentation.

The perspective from which the maize is viewed plays a significant role in
detection accuracy.
Most of the data were captured from a nadir view (top-down) using
low-altitude \uav flights.
However, because of camera parallax and angles, the resulting images
include some perspective distortion.
In nadir view, plants located near the edges of the frame often appear
slanted due to perspective distortion and lens projection geometry.
Although the camera is oriented directly downward, objects near the center
of the image are captured from a true top-down perspective, while those
closer to the edges are viewed at steeper angles.
As a result, plants near the right edge appear to lean to the right, those
on the left to the left, and those near the top or bottom slant upward or
downward, respectively.
This distortion causes the same plant to present very different visual cues
depending on its position in the frame, making consistent detection more
difficult.
When images are captured from an oblique angle, an additional challenge
arises -- plants closer to the camera, typically near the bottom of the
frame, appear larger, while those farther away, near the top, appear
smaller.
This introduces significant within-frame scale variation with which models not
trained on such diversity may struggle (see Panel \ref{obs}).
%
To address this, we fragmented the frames into smaller dimensions to avoid
downsampling before feeding the images to the model, such as using \(640
\times 640\)
instead of \(1920 \times 1920\), allowing the plants to retain higher resolution.
While this helped in some cases, it still affected the evaluation of
non-nadir views.
As shown in the evaluation results in Table \ref{general},
frames with
oblique angles still scored lower than those captured with nadir views.
It is also important to recognize that evaluation metrics like \map, while
useful, do not always reflect practical performance.
In agricultural tasks such as stand counting, the primary goal is often to
determine the number of individual plants in an area, rather than to
perfectly classify each instance.
A model that predicts two singles instead of one double
%
%
can still produce
an accurate count.
For this reason, alternative approaches -- such as lowering the confidence
threshold followed by post-processing -- can improve counting precision.
This can be followed by post-processing steps, such as removing duplicates
based on confidence scores or prioritizing higher or lower class
predictions, depending on whether overcounting or undercounting is
preferred.
Together, these strategies can help improve the precision of the final
results.
In addition to accuracy, inference efficiency is a critical factor for
deploying seedling detection models in the field. Table \ref{exec-time}
shows the execution time for each model across 806 test images, providing a
clear comparison of their processing speed.
\yoloe, in particular, stands out as the fastest model, completing
evaluation in just 29 seconds and full detection in 125 seconds. This
corresponds to approximately 27.7 and 6.4 frames per second (fps),
respectively.
This is well within the range for near real-time processing on edge devices
or onboard computing systems in tractors and \uavs.
\yolot and \yolon also demonstrate competitive performance, taking 165 and
243 seconds respectively for detection, offering viable options for
embedded applications where some latency is acceptable.
In contrast, \frcnn is significantly slower, requiring 684 seconds for
detection, which limits its practicality for real-time field use without
high-performance compute resources.
These results show that \yolo-based models, especially \yoloe, are not only
accurate but fast enough to support on-the-fly seedling detection, enabling
integration into existing agricultural machinery for real-time monitoring
and decision-making.

\section{Conclusion}
\label{sec:orge95fcc8}

This paper presents the capabilities and limitations of current object
detection models for maize seedling detection under realistic field
conditions.
All evaluated models, including \yolon, \yoloe, \yolot, and
\frcnn, performed reliably when detecting single seedlings, particularly
during the V4--V6 growth stages and in nadir-view images.
However, detection accuracy decreased notably for double and triple
clusters of seedlings due to class imbalance, annotation ambiguity, and occlusion in
dense plantings.
The \smd dataset, designed to be diverse in genotypes,
lighting conditions, soil types, and camera angles, improved generalization
but also introduced challenges.
Oblique views and perspective distortion reduced detection consistency,
emphasizing the importance of image framing and resolution during both
training and inference.
Synthetic data generation for rare classes provided limited improvements,
suggesting a need for more realistic augmentation methods such as
generative adversarial networks (GANs) or 2.5D--3D simulation, which
could help models better learn complex and occluded plant structures.
While \map remains a valuable metric, it does not fully
capture the practical requirements of stand counting, where approximate
counts may suffice.
Post-processing strategies like confidence filtering
and duplicate suppression can help tailor model outputs to specific
agricultural applications.
Inference speed is also a critical factor for
field deployment; \yoloe achieved over 27 fps, enabling near real-time
performance, whereas \frcnn was significantly slower, limiting its
practicality.
This highlights the suitability of lightweight \yolo-based
models for onboard or edge inference in \uav or tractor-mounted
systems.
Future work should aim to enhance the dataset with more diverse plant
groupings, increase coverage of oblique-view imagery, and incorporate
instance segmentation to improve detection accuracy under complex field
conditions.
These advancements would further support the scalability and
robustness of \uav-based crop monitoring systems, enabling a wide range of
downstream applications such as variable-rate planting, selective
replanting, early weed pressure detection, and data-driven yield
forecasting. Collectively, these improvements bring precision agriculture
closer to full-scale field deployment.

\subsection*{Acknowledgments}
\label{sec:orgce13089}
We extend our sincere gratitude to our colleagues at the Missouri
Maize Center, with special thanks to our farm manager, Chris Browne.
The success of our imaging depends heavily on outstanding weed control,
and this work included imaging in several other fields in addition to our
own.
We also thank Hadi AliAkbarpour, Filiz Bunyak, Kannappan Palaniappan,
Matthew Stanley, Dexa Akbar, Rifki Akbar, Bill Wise, and Vinny Kazic-Wise
for their valuable discussions and insights.

\subsection*{Conflict of Interest Statement}
\label{sec:org5b33db6}
The authors declare that the research was conducted in the absence of any
commercial or financial relationships that could be construed as a
potential conflict of interest.

\subsection*{ORCID}
\label{sec:org8891930}
\begin{itemize}
\item Dewi E. Kharismawati  \url{https://orcid.org/0000-0002-3063-1618}

\item Toni Kazic \url{https://orcid.org/0000-0001-5971-2406}
\end{itemize}

\subsection*{Data Availability Statement}
\label{sec:org844ae7a}

The \smd dataset is publicly available in the \href{https://drive.google.com/drive/folders/1EC2aR1HbCsRnIJbXWeEHddyGTQpGFh2n?usp=sharing}{\smd dataset}.
The directory structure for both \yolo and \frcnn is shown in Figure
\ref{data-structure}.
The source code for data labeling, training, validation, and testing can be
found in the \href{https://github.com/dek8v5/MaizeSeedlingDetectionDataset.git}{\smd Repository}.

\subsection*{Author Contributions}
\label{sec:org98e80f3}
Both authors contributed to the conceptualization, methodology, validation,
resources, and writing (draft and revisions).  D.E.K. was responsible for
software development, data collection and curation, implementation and
testing, and visualization.  T.K. provided supervision, project
administration, and funding acquisition.  Both authors have read and agreed
to the published version of the manuscript.

\subsection*{Funding}
\label{sec:orge670b49}
We gratefully acknowledge support from the Dept. of
Electrical Engineering and Computer Science for D.E.K. and an anonymous
gift in aid of maize research.



\bibliographystyle{IEEEtran}
\def\localdots{../../..}

%
%

%
%


\def\db#1{\localdots/bibliography/#1.bib}




%



%
 
\def\bp{\db}

%

\bibliography{\bp{journals},%
              \bp{keys},%
              \bp{miscellaneous},%
              \bp{clean-egbib},%
              \bp{nascent},%
              \bp{all}}


\begin{IEEEbiography}[
{\includegraphics[width=1in,height=1.25in,clip,keepaspectratio]{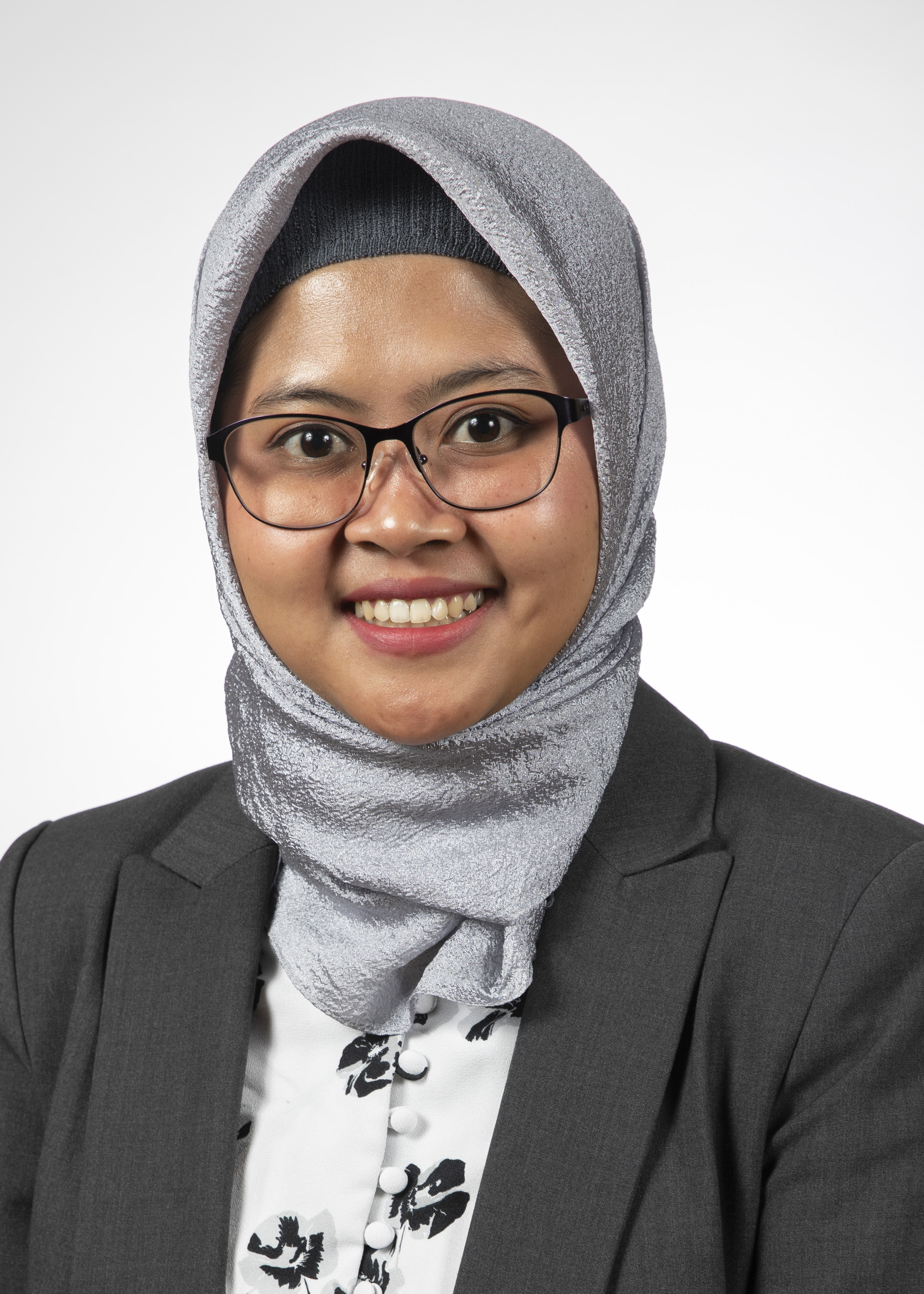}}]
{Dewi Endah Kharismawati} (Member, IEEE)
received the B.S. and Ph.D. degrees in Computer Science from the University of Missouri, Missouri, USA,
in 2017 and 2025, respectively. She has been a Postdoctoral Researcher with the AgSensing Lab, Department
of Food, Agricultural, and Biological Engineering, The Ohio State University, since June 2025.
Her research interests include computer vision, deep learning, and image processing applied to
UAV aerial imagery, with a focus on image registration, plant detection, and 3D reconstruction for
generating actionable insights for farmers and agricultural researchers.
\end{IEEEbiography}
\begin{IEEEbiography}[
{\includegraphics[width=1in,height=1.25in,clip,keepaspectratio]{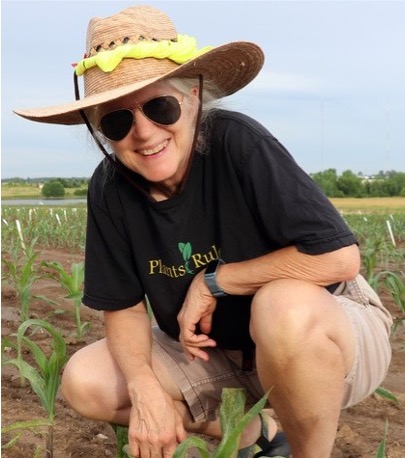}}]
{Toni Kazic} received the B.S. degree in microbiology from the University of Illinois,
Urbana, in 1975, the Ph.D. degree in genetics from the University of Pennsylvania in 1984,
and was a postdoctoral fellow in bacterial genetics at the Institute for Cancer Research,
Fox Chase Cancer Center, Philadelphia, and Washington University, St. Louis before switching
to computational biology.
Since 2006, her work has focused on computational and genetic studies of the
biochemical networks underlying complex phenotypes in maize. She is an associate
professor in the Department of Electrical Engineering and Computer Science at the
University of Missouri, Columbia. Her lab, the Missouri Maize Computation and Vision
(MMCV) Lab, investigates a family of maize mutants that exhibit lesion phenotypes,
which are visible spots on leaves and define a high-dimensional phenotypic manifold.
To scale up field experiments for higher resolution, better sample size, and more robust
quantitation, her group combines high throughput phenotyping with consumer-grade drones
with developing computational methods to monitor plant growth and lesion formation.
She has been a fellow at the NIH (Division of Computer Research and Technology),
Argonne National Laboratory (Division of Mathematics and Computer Science), and ICOT,
the Japanese Institute for Fifth Generation Computer Technology.
She has served as a Program Director at the National Science Foundation in
computational biology, as board member and secretary of the International Society
of Computational Biology, and as an associate member of the Joint Nomenclature Commission
of the International Union of Pure and Applied Chemistry and the International Union of
Biochemistry and Molecular Biology (the ``Enzyme Commission'').
She was elected a fellow of the American College of Medical Informatics in 2004.
\end{IEEEbiography}

\end{document}